\documentclass{ieeeaccess}
\usepackage{cite}
\usepackage{algorithmic}
\usepackage{graphicx}
\usepackage{latexsym}
\usepackage{textcomp}
\usepackage{amsfonts,amssymb,amsmath,amsthm}
\usepackage{multirow}
\usepackage{multicol}
\usepackage{hhline}
\usepackage{hyperref}
\usepackage[english]{babel}
\usepackage{makecell}

\theoremstyle{definition}
\newtheorem{definition}{Definition}[section]

\def\BibTeX{{\rm B\kern-.05em{\sc i\kern-.025em b}\kern-.08em
    T\kern-.1667em\lower.7ex\hbox{E}\kern-.125emX}}
\begin{document}

\history{Date of publication xxxx 00, 0000, date of current version xxxx 00, 0000.}
\doi{10.1109/ACCESS.2017.DOI}

\title{A Baselined Gated Attention Recurrent Network for Request Prediction in Ridesharing}

%-----------------
\author{\uppercase{Jingran Shen\authorrefmark{1}},
\uppercase{Nikos Tziritas\authorrefmark{2}},
\uppercase{Georgios Theodoropoulos\authorrefmark{3}}.
}
\address[1]{Dept. of Computer Science and Engineering, Southern University of Science and Technology (SUSTech), Shenzhen, China (e-mail:petershen815@126.com)}
\address[2]{Department of Informatics and Telecommunications, University of Thessaly, Greece (e-mail: nitzirit@uth.gr)}
\address[3]{Dept. of Computer Science and Engineering, and  Research Institute for Trustworthy Autonomous Systems, Southern University of Science and Technology (SUSTech), Shenzhen, China  (e-mail: theogeorgios@gmail.com)}
\tfootnote{This research was supported by: Shenzhen Science and Technology Program,  China (No. GJHZ20210705141807022); Guangdong Province Innovative and Entrepreneurial Team Programme, China (No. 2017ZT07X386); SUSTech Research Institute for Trustworthy Autonomous Systems, China.}

\markboth
{Shen, Tziritas  and Theodoropoulos: A Baselined Gated Attention
Recurrent Network for Request
Prediction in Ridesharing}
{Shen, Tziritas  and Theodoropoulos: A Baselined Gated Attention
Recurrent Network for Request
Prediction in Ridesharing}

\corresp{Corresponding authors: Jingran Shen (e-mail: petershen815@126.com) and Georgios Theodoropoulos (e-mail: theogeorgios@gmail.com).}

%%% Abstract %%%
\begin{abstract}
Ridesharing has received global popularity due to its convenience and cost efficiency for both drivers and passengers and its strong potential to contribute to the implementation of the UN Sustainable Development Goals. 
%With the design of grouping multiple passengers from different requests, the drivers are able to reduce their travel costs while in the meantime, the passengers are able to gain better travel experiences with other passengers. 
As a result, recent years have witnessed an explosion of research interest in the \textit{RSODP} (Origin-Destination Prediction for Ridesharing) problem with the goal of predicting the future ridesharing requests and providing schedules for vehicles ahead of time. Most of the existing prediction models utilise Deep Learning. However, they fail to effectively consider both spatial and temporal dynamics. In this paper the Baselined Gated Attention Recurrent Network (\textit{BGARN}), is proposed, which uses graph convolution with multi-head gated attention to extract spatial features, a recurrent module to extract temporal features, and a baselined transferring layer to calculate the final results. The model is implemented with PyTorch and DGL (Deep Graph Library) and is experimentally evaluated using the New York Taxi Demand Dataset. The results show that 
%the RMSE, MAPE and MAE for 
BGARN outperforms all the other existing models in terms of prediction accuracy.
\end{abstract}

%%% Keywords %%%

\titlepgskip=-15pt

\maketitle

%%% Main-matter %%%
\section{Introduction}
\label{sec:intro}

%\subsection{Ridesharing with OD Prediction}

\PARstart{R}{idesharing} is an increasingly popular service paradigm where passengers from different requests are grouped into shared rides in order to achieve certain objectives, such as reduced travel expenses for vehicle drivers and improved vehicle dispatching flexibility for the service platform. With the development of GIS services and online ride-hailing applications like Didi, Lyft and Uber, there is an abundance of the road network data and passenger request data that underpin the development of dispatching algorithms \cite{Mitropoulos}. 

\begin{figure}
    \centerline{\includegraphics[width=3.5in]{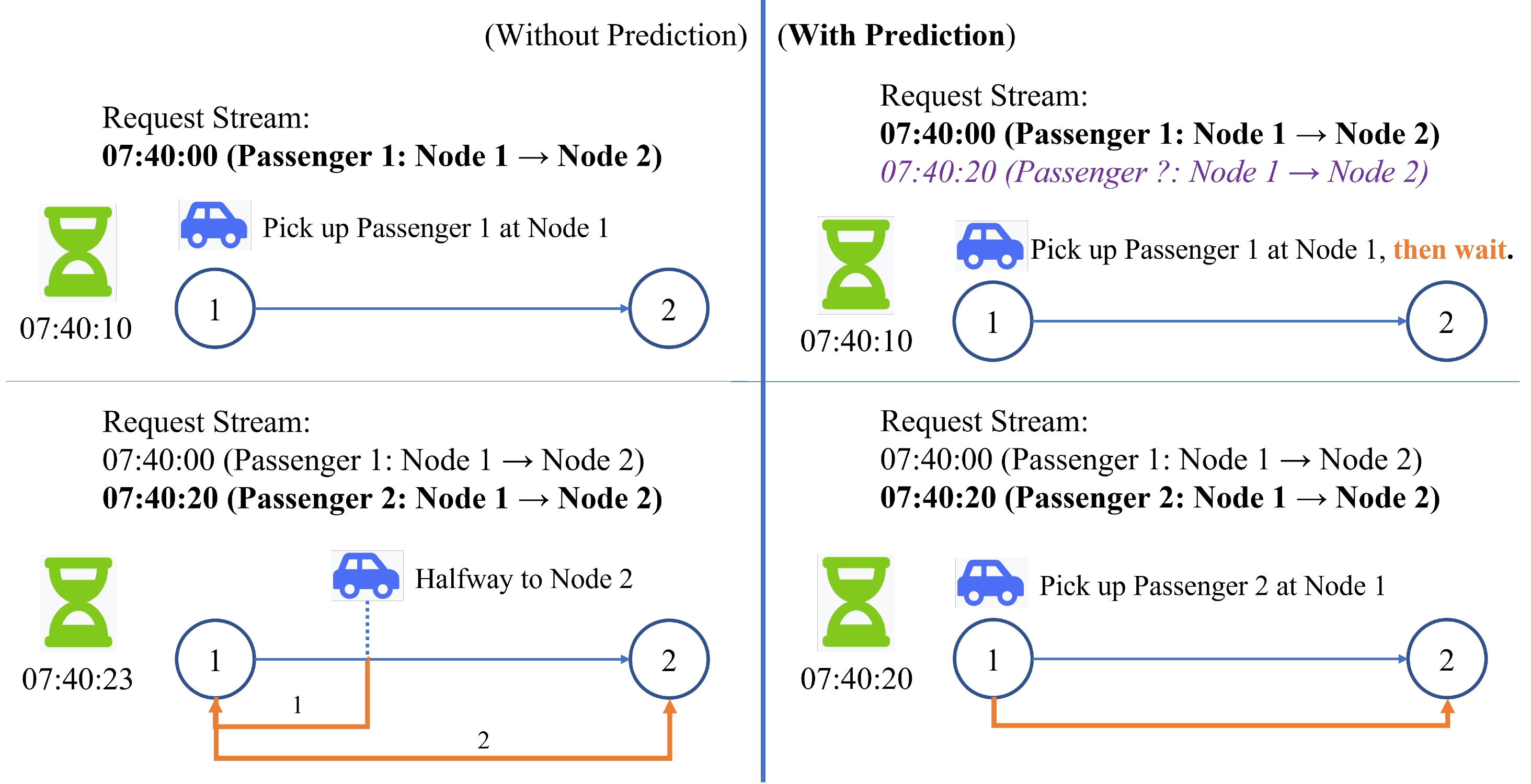}}
    \caption{Vehicle Schedule Optimization with Prediction. The figure shows two optimization results of two timestamps with/without prediction in advance. Suppose the requests are fed into the system every 20 seconds. In this example, each of the two latest request batches contains one request with the same OD, but different passenger ID. Without prediction, the vehicle will immediately start to handle the request from passenger 1. When passenger 2 submits the other request in the next 20 seconds, the vehicle is already halfway from node 1 to node 2 so it needs to re-route to handle the request from passenger 2, which incurs extra travel cost. With prediction of the future request in the first 20 seconds, the vehicle will be arranged to simply wait 10 seconds for the predicted request.}
    \label{fig:prediction_helps}
\end{figure}

For the past few years, a tremendous amount of research has been conducted on traffic forecasting \cite{2021arXiv210111174J} with the aim of tackling problems related to topics like transportation planning and environmental protection. With regards to ridesharing, one significant case is \textit{RSODP} (Origin-Destination Prediction for Ridesharing, also referred to as \textit{OD prediction}), which aims to foresee the pattern of passenger requests in the future for improved request-vehicle assignment. Figure \ref{fig:prediction_helps} shows an example of how OD prediction aids the vehicle schedule planning in real time. It is clear to notice that with the prediction request produced 20 seconds ahead of time, the vehicle is scheduled to wait at node 1 instead of starting its delivery of passenger 1 immediately to node 2. As a result, the travel cost of the vehicle is minimized. Another simple case is that the vehicle travels to node 1 when idle from delivery at 6:00 p.m. because prediction indicates that a considerable number of requests will be submitted there before long. 

Predicting the OD requests means not only the practicability to provide vehicle schedules in advance, but also the possibility to maintain passenger mobility. Understanding how people move around the city has potential benefits to other decision making tasks such as passenger travel behavior analysis. Intuitively, a chauffeur/chauffeuse who offers customized driving services to a certain passenger, needs to learn the travel behavior of the passenger (e.g., the passenger regularly requests a trip from node 1 to node 2 at 6:00 p.m.) in advance so that he/she can plan the optimal driving route with extra considerations including traffic congestion as well as the weather.

 {} 

Contributing to the effort to address the RSODP problem, this paper proposes a new model, referred to as BGARN, Baselined Gated Attention Recurrent Network. The model utilises graph convolution with multi-head gated attention to extract spatial features, a recurrent module to extract temporal features, and a baselined transferring layer to calculate the final results. 

The contributions of the paper are as follows:
\begin{enumerate}
    \item It introduces a new model, referred to as BGARN, for addressing the RSODP challenge.
    \item It utilises multi-head gated attention to combine different perspectives describing the semantic relationship among geographical grids.
    \item It proposes a hybrid approach, termed in this paper as \textit{tuning}, which combines linear baseline results with non-linear neural network results to enhance the predictive capability of the model. Different combination approaches are tested and discussed.
     \item It presents a detailed comparative experimental analysis with existing state of the art models using the New York Yellow Taxi Trip dataset.
\end{enumerate}

The remainder of the paper is organised as follows. Section \ref{sec:related} provides an outline of existing RSODP algorithms while section \ref{sec:preliminaries} presents an explicit definition of the RSODP problem. Section \ref{sec:model} provides a comprehensive explanation of the proposed BGARN model. Section \ref{sec:experiment} presents a comparative quantitative experimental analysis of the proposed system. Section \ref{sec:conclusion} summarizes the paper and outlines future research directions. 

\section{Related Work}
\label{sec:related}

\begin{figure}
    \centerline{\includegraphics[width=3.5in]{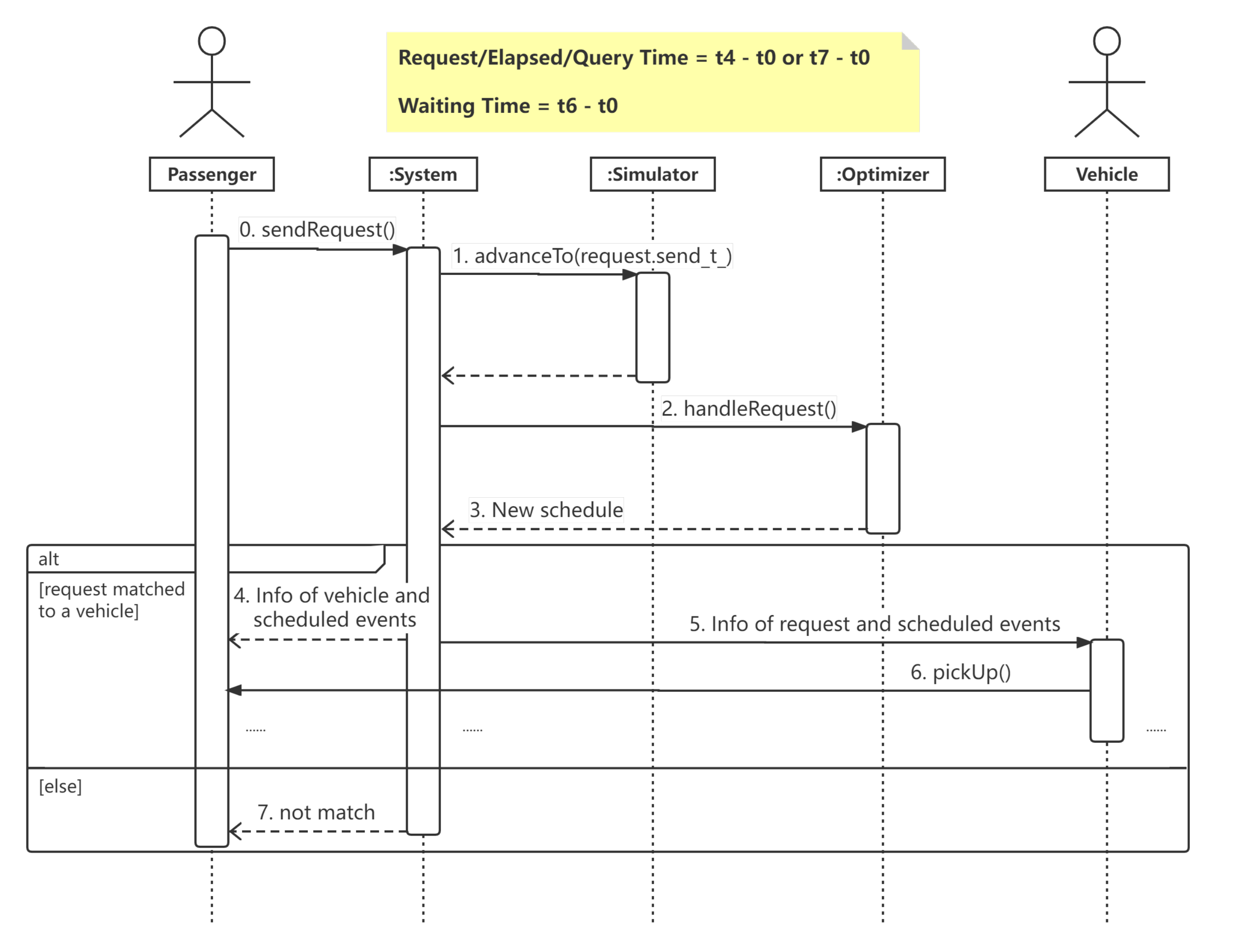}}
    \caption{General Ridesharing Process (adapted from \cite{10.14778/3339490.3339493})}
    \label{fig:general_ridesharing_process}
\end{figure}

A general ridesharing process includes a central system, a simulator and an optimizer, as illustrated in Figure \ref{fig:general_ridesharing_process}. Such a system is responsible for passing the incoming requests from passengers to the optimizer and then dispatching the requests to the vehicles according to the returned schedule. The optimizer may utilise different ridesharing algorithms in order to provide the best request-vehicle assignment plan. The simulator is used to support algorithm selection since it evaluates the request handling metrics in the future.

Several OD prediction models have been developed. LSTNet \cite{lai2018modeling}, GCRN \cite{Seo2018StructuredSM}, GEML \cite{10.1145/3292500.3330877} and Gallat \cite{Wang2021PassengerMP} represent the most influential state of the art systems with Gallat reported to outperform the rest \cite{Wang2021PassengerMP}. 

The majority of existing OD prediction models utilize Deep Learning to capture the spatial as well as temporal features of the OD requests \cite{Wang2021PassengerMP, Jin2020DeepMS, Ke2019PredictingOR, 10.1145/3292500.3330877, Xu2019IncorporatingGA, 8566163, 9101359, Wang2020AnUC, Pian2020SpatialTemporalDG, lai2018modeling, Seo2018StructuredSM}. Although there have been attempts to develop statistical models such as \cite{10.1145/2020408.2020579}, the prediction accuracy turns out to be unsatisfactory (lower than 50\%). One possible explanation could be that the complex spatial-temporal dynamics of the passenger request stream on the city-wide road network cannot be captured by simply stacking Gaussian distributions with limited parameters.
 
The Deep Learning models proposed in the past few years tend to choose GNN (Graph Neural Network)  over CNN (Convolutional Neural Network) to capture the non-Euclidean spatial dynamics and a recurrent module such as LSTM (Long Short-Term Memory) to capture the temporal dynamics. Typically, the request data is preprocessed into a grid map in which grids represent the geographical regions. Such grid map is also referred to as an OD Graph since each edge represents the number of requests from one grid to another. Aggregated with the spatial features such as Semantic Neighborhood and haversine distance, the grids from the OD Graph form a new feature graph in which one directed edge represents the propagation tendency of the features from one grid to another. With the support of GNN, the feature propagation among the nodes is performed and the output grid matrix specifies the extracted request forwarding pattern. A time series of these grid matrices is then pushed into a recurrent module like LSTM to integrate the temporal features such as tendency and periodicity (also referred to as trend and period explained in \cite{10.5555/3298239.3298479}). Finally, the output spatial-temporal features are translated back to a new OD Graph in order to predict the future request flow.

GCRN \cite{Seo2018StructuredSM} does not utilize the spatial features. Instead, it processes the OD graph directly as a spatial feature, which severely increases the initial feature dimension. In order to resolve this issue, later proposed models like GEML \cite{10.1145/3292500.3330877} and Gallat \cite{Wang2021PassengerMP} specify Semantic Neighborhood to examine the existence of requests from one grid to another and Geographical Neighborhood to examine the haversine distance between two grids. Gallat further splits Semantic Neighborhood into Forward and Backward Neighborhood to stress the importance of distinguishing the two roles of the grids as the origin or destination. Additionally, GEML and Gallat utilise pre-weights which take the contributions of different neighbors into account. This technique allows neighbors with more intensive request flow to contribute more to the feature extraction operation.

One of the deficiencies of applying simple GNN is that the importance of feature propagation along different edges should be different (the edge weights are unequal) since the affinities among grids are different due to the spatial dynamics. To address this issue, the Attention mechanism \cite{10.5555/3295222.3295349} was proposed in 2017, and, subsequently, the GAT (Graph Attention Network) model \cite{velickovic2018graph} incorporated attention into GNN and turned out to be a feasible solution to leveraging the edge weights.  One of the currently state-of-the-art models, Gallat (Graph prediction with all attention) \cite{Wang2021PassengerMP}, utilized a pre-weighted GAT to distinguish the edge weights. In 2018, a gated-GAT called GaAN (Gated Attention Network) \cite{Zhang2018GaANGA} was proposed to handle the traffic speed forecasting problem. The model introduced importance to each attention head \cite{10.5555/3295222.3295349} by using a novel concept referred to as gates \cite{Zhang2018GaANGA}, which provides a potential upgrade perspective for Gallat.

From the temporal perspective, \cite{Xu2019IncorporatingGA} replaced LSTM with GRU (Gated Recurrent Unit) to reduce computation cost while \cite{Jin2020DeepMS} used a Transformer from \cite{10.5555/3295222.3295349} to capture the long-term temporal dynamics. The Transformer parallelises the computations of time series by maintaining the semantic embeddings among the input sequence units. Gallat also refers to this design and modifies the temporal processing unit according to the self-attention concept extracted from the Transformer. Nevertheless, such technique serves more as a supplemental spatial extraction, and suffers the risk of losing information which can be inferred by the continuity of time.

Finally, the aforementioned structure contains three modules responsible for three different tasks, namely spatial feature extraction, temporal feature extraction as well as feature translation which produces the final prediction results. Such complex model structure might easily suffer from gradient explosion or vanishing if weights are carelessly processed. Apart from using residual block and normalization, a simple baseline model can also be used to provide reference outputs. The baseline outputs provide a rough estimate which the deep learning results can improve upon. As an example, LSTNet \cite{lai2018modeling}
combines the prediction outputs generated by the deep neural network and those generated by a baseline AR (Auto-Regressive) model using addition.

\subsection{Towards a new model}

Based on the above discussion, it is evident that there are several aspects for a good solution to the problem that existing systems do not all cover. For the preprocessing stage, the grids can be partitioned into hexagons rather than rectangles as suggested in \cite{8566163}, since hexagons have smaller perimeter-to-area ratio as well as higher isotropy; for the spatial layer, multi-head gated attention can be introduced to investigate the spatial feature space on a finer granularity; for the temporal layer, as explained above, sticking to the conventional recurrent module design is a more viable option. Finally, as suggested by \cite{lai2018modeling}, baseline results from simple linear models may be used as a reference in the final prediction layer to enhance the predictive capability of the models.

BGARN, introduced in this paper, aims to address these gaps. 
Table \ref{tab:model_comparison} summarises and contrasts the features supported and utilised  by the four state of the art models, LSTNet, GCRN, GEML and Gallat, and  the proposed BGARN model.

\setlength{\arrayrulewidth}{1pt}
\setlength{\tabcolsep}{3.2pt}
\renewcommand{\arraystretch}{2.0}
\begin{table}
    \caption{Model Comparison}
    \label{tab:model_comparison}
    \centering
    \begin{tabular}{|c | c|c|c|c | c|}
        \hline
         Features & LSTNet & GCRN & GEML & Gallat & \textbf{BGARN} \\
        \hline
        \makecell{Process OD \\ Graph directly} & 
        \checkmark & \checkmark &  &  &  \\
        \hline
        \makecell{Use GNN} & 
         & \checkmark & \checkmark & \checkmark & \checkmark \\
        \hline
        \makecell{Use Attention} & 
        \checkmark &  & \checkmark & \checkmark & \checkmark \\
        \hline
        \makecell{Multi-head \\ Gated Attention} & 
         &  &  &  & \checkmark \\
        \hline
        \makecell{Forward \& Backward \\ Neighborhood} & 
         & \checkmark &  & \checkmark & \checkmark \\
        \hline
        \makecell{Geographical \\ Neighborhood} & 
         & \checkmark & \checkmark & \checkmark & \checkmark \\
        \hline
        \makecell{Use pre-weights} & 
         &  & \checkmark & \checkmark & \checkmark \\
        \hline
        \makecell{Use Recurrent \\ Module} & 
        \checkmark & \checkmark & \checkmark &  & \checkmark \\
        \hline
        \makecell{Examine Tendency} & 
        \checkmark & \checkmark &  & \checkmark & \checkmark \\
        \hline
        \makecell{Examine Periodicity} & 
         &  & \checkmark & \checkmark & \checkmark \\
        \hline
        \makecell{Use Baseline} & 
        \checkmark &  &  &  & \checkmark \\
        \hline
    \end{tabular}
\end{table}

%-----------------------------------------
\section{Problem Definition}
\label{sec:preliminaries}
%-----------------------------------------

In this section, RSODP and important concepts are mathematically defined.

\begin{definition}[Time Slot]
\label{def:timesep}

A time slot $t \in [1, T], t \in \mathbb{N}^+$ represents the minimum time unit for handling the requests. Each time slot is of $t_n$ time endurance in hours. By separating the requests every $t_n$ hours, $T$ time slots of requests are generated in total. For RSODP, $t_n = 1$ seems to be a reasonable time endurance to fully process the request data and have the model predict the requests in the next hour.

\end{definition}

\begin{figure}
    \centerline{\includegraphics[width=3.5in]{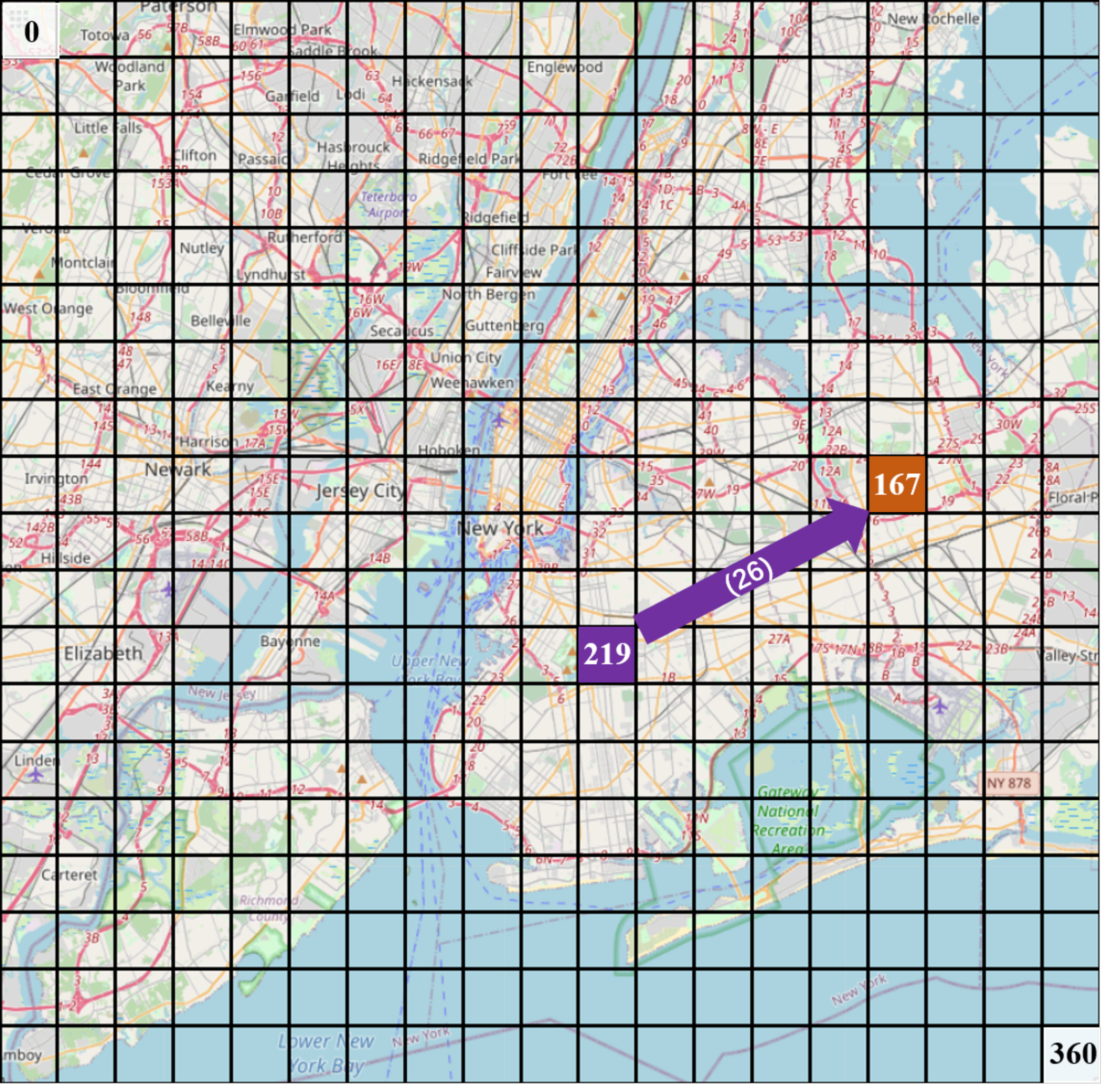}}
    \caption{Grid Map of New York City. Separated into $19 \times 19$ grids according to the longitudes and latitudes. Suitable grid size is around $2.6$ km $\times$ $2.6$ km - a $30$ km/h vehicle trip of 5 minutes. In this example, the number of requests at this time slot from grid 219 to grid 167 is 26 (value in parenthesis).}
    \label{fig:grid_map}
\end{figure}

\begin{definition}[Grid]
\label{def:grid}

A map of city is partitioned into several grids according to the longitudes and latitudes such that they cover the region of the city without intersecting one another. As implemented in \cite{Jin2020DeepMS, 10.1145/3292500.3330877, Wang2021PassengerMP, Wang2020AnUC, Xu2019IncorporatingGA, 9101359, Pian2020SpatialTemporalDG}, the grids are partitioned into rectangular shape for easier computations. In real world case, each grid is approximately of $2.6$ km $\times$ $2.6$ km size. Figure \ref{fig:grid_map} shows an example of partitioning New York City \footnote{The raw map without grids and labels is cropped from \href{https://www.openstreetmap.org/}{OpenStreetMap}.} into a grid map. There is, though, another grid partition technique suggested in \cite{8566163} that produces hexagon grids to better describe the affinity between a grid and its geographical neighbors. This technique can be attempted on in the future.

\end{definition}

\begin{definition}[Origin-Destination Graph]
\label{def:odgraph}

An OD Graph $G_t = (\mathcal{V}_t, \mathcal{E}_t, R) \in \mathbb{N}^{n \times n}$ at time slot $t$ is a snapshot graph representing the origin-destination relationships among the grids. Each of the $n$ grids is considered as a node $v_i^t \in \mathcal{V}_t$ where $i$ represents the grid ID in the graph. A directed weighted edge $e_{i, j}^t = (v_i^t \rightarrow v_j^t, g_{i, j}^t) \in \mathcal{E}_t$ represents the request flow from grid $i$ to grid $j$, with a total number of $g_{i, j}^t$ requests appearing. Again, Figure \ref{fig:grid_map} shows an example of an edge in an OD graph. In this case, the number of requests starting from grid 219 to grid 167 is 26. Regardless of time slots, the geographical adjacency matrix $R$, in which each $r_{i, j} \in R$ denotes the haversine distance between the central coordinates of grid $i$ and grid $j$, remains unchanged. $\{G_t\}_{t=1}^T = \{G_1, G_2, ..., G_T\}$ represents all $T$ OD graphs in the input sequence.

\end{definition}

\begin{definition}[Request]
\label{def:req}

A request $d = (t_r, \allowbreak \text{lat}_o, \allowbreak \text{lng}_o, \allowbreak \text{lat}_d, \allowbreak \text{lng}_d) \in \mathcal{D}$ stores the request time $t_r$ as well as the coordinates of the origin and destination as a tuple of latitude and longitude $(\text{lat}_o, \text{lng}_o)$, $(\text{lat}_d, \text{lng}_d)$. It is preprocessed to construct the OD graph $G_t$.

\end{definition}

\begin{definition}[\textbf{RSODP: Origin-Destination Prediction for Ridesharing}]
\label{def:rsodp}

Given a sequence of requests $\mathcal{D}$ which is later transformed into a sequence of $T$ OD graphs $\{G_t\}_{t=1}^T$, the geographical adjacency matrix $R$ and basic grid map information, RSODP aims to predict $G_{T+1}$, the OD graph in the next time slot in the future.

\end{definition}

\section{System Architecture}
\label{sec:model}

\begin{figure}
    \centerline{\includegraphics[width=3.5in]{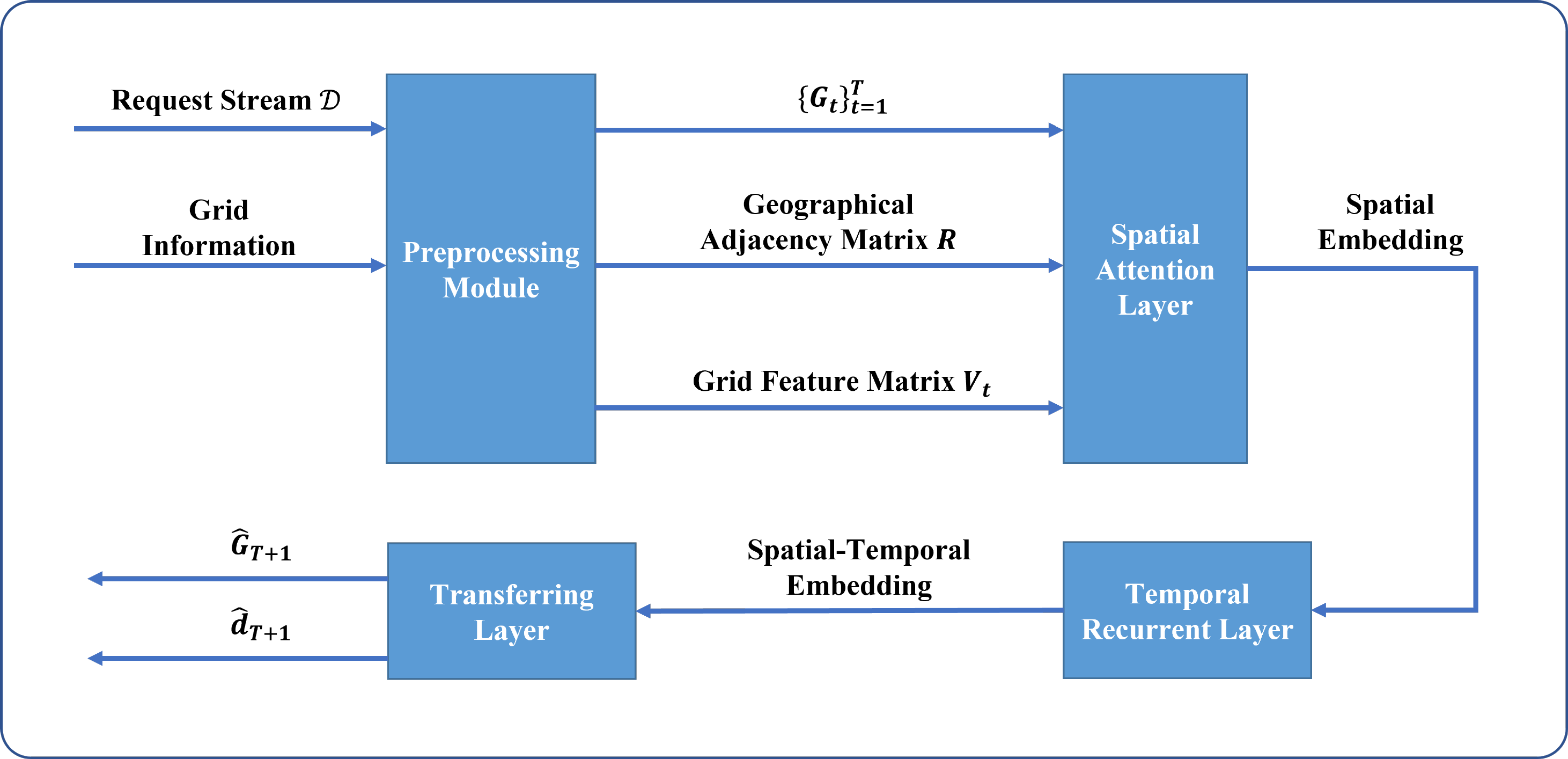}}
    \caption{The architecture of BGARN. Raw data will go through four key components, namely the preprocessing module, spatial attention layer, temporal recurrent layer as well as the transferring layer. The model outputs a demand vector $\hat{d}_{T+1}$ and the predicted OD graph $\hat{G}_{T+1}$ for the next time slot in the future.}
    \label{fig:model}
\end{figure}

This section provides a detailed description of the proposed BGARN system. The overall structure of BGARN is depicted in Figure \ref{fig:model}. The input raw data is the request stream $\mathcal{D}$ and grid information specifying the boundaries of the city on the map, grid size as well as the number of grids on latitude and longitude directions. The preprocessing module transforms the requests into the OD graph sequence $\{G_t\}_{t=1}^T$, generates geographical adjacency matrix $R$ and a grid feature matrix $V_t \in \mathbb{R}^{n \times d}$ from the grid information. All these outputs are then passed into the spatial attention layer and temporal recurrent layer sequentially. As a result, the spatial-temporal embeddings will be computed to store the features of the affinities among grids. Eventually, the embeddings are passed through a transferring layer to translate the features into two required outputs: a demand vector $\hat{d}_{T+1} \in \mathbb{R}^{n}$ (as a subtask) storing the predicted number of outgoing requests from each grid, and the predicted OD graph $\hat{G}_{T+1} \in \mathbb{R}^{n \times n}$.

\subsection{Preprocessing Module}

There are basically three tasks to handle in the preprocessing module.

First, a grid map is generated using the grid information including the boundaries of the region of interest (e.g., a city) and the specified grid size for reference. Starting with 0 from the top left, each grid is labeled an ID from left to right and top down. Subsequently, the haversine distances among the grids are calculated to form up the geographical adjacency matrix $R$.

Next, the OD graph sequence $\{G_t\}_{t=1}^T$ is generated. Each request is mapped to a specific time slot $t$ using the request time $t_r$. The coordinates of the origin and destination are mapped to two grids $i$ and $j$. And $1$ is added to the corresponding $g_{i, j}^t$.

Finally, a grid feature matrix $V_t$ is created using the OD graph $G_t$ and the grid information. The feature of a grid can vary through multiple dimensions. For example, the coordinates and ID of the grid represent its geographical properties. The day of week (e.g., Monday as 0 and Sunday as 6), hour of day represent the time-related information (BGARN also uses weekday/weekend and time period such as "afternoon" in the implementation). The out-degree and in-degree of the grid represent the semantic features. Additionally, there are several auxiliary features including the functionality of the grid (e.g., residential area or workplace), weather for the time, etc. The feature vector $\mathrm{v}_i^t \in \mathbb{R}^{d_f}$ of grid $i$ at time slot $t$ can thus be constructed by concatenating all the transformed features together.

\subsection{Spatial Attention Layer}

The spatial attention layer, which was introduced by \cite{Wang2021PassengerMP}, intends to extract spatial features from the grids and form up the spatial affinities among them. With limited information of the grids provided, the model specifies three views of spatial affinity analysis: \textit{Forward Neighborhood}, \textit{Backward Neighborhood} as well as \textit{Geographical Neighborhood}. 
For completeness and convenience of the reader and to enable better understanding of the proposed model, the formal definitions of the aforementioned concepts as described in the original papers \cite{10.1145/3292500.3330877,Wang2021PassengerMP} are provided below (Equations \ref{eq:fwdNbh}-\ref{eq:spa_embed}).

\begin{definition}[Forward Neighborhood]
\label{def:fwdNbh}

If there is at least one request from grid $i$ to grid $j$, then grid $j$ is a forward neighbor of grid $i$. The set of forward neighbors for grid $i$ at time slot $t$ can be defined as follows \cite{Wang2021PassengerMP}:
\begin{equation}
\label{eq:fwdNbh}
\Psi_i^t = \{j | g_{i, j}^t > 0, g_{i, j}^t \in G_t\}.
\end{equation}

\end{definition}

\begin{definition}[Backward Neighborhood]
\label{def:bwdNbh}

Correspondingly, if there is at least one request from grid $j$ to grid $i$, then grid $j$ is a backward neighbor of grid $i$. The set of backward neighbors for grid $i$ at time slot $t$ can be defined as follows \cite{Wang2021PassengerMP}:
\begin{equation}
\label{eq:bwdNbh}
\Phi_i^t = \{j | g_{j, i}^t > 0, g_{j, i}^t \in G_t\}.
\end{equation}

\end{definition}

Intuitively, the OD neighborhood specification tends to describe the tendency of people flowing from one grid to the other. If there are frequent requests happening between two grids, then future requests might have a higher probability to take place between the two as well. By aggregating the features of these neighbor grids together, the mobility pattern in the region can be effectively examined. The importance to consider forward and backward neighborhood separately, since they are distributed in time quite differently, is explained in \cite{Wang2021PassengerMP}, while \cite{10.5555/3298239.3298479} has shown that the propagation from a grid to its forward neighbors can be affected by that of its backward neighbors (e.g., continuous commuting to work).

\begin{definition}[Geographical Neighborhood]
\label{def:geoNbh}

If the haversine distance between two grids is within a specified threshold $L$, then grid $i$ and grid $j$ are considered as geographical neighbors of each other. The set of geographical neighbors for grid $i$ can be defined as follows \cite{10.1145/3292500.3330877,Wang2021PassengerMP}:
\begin{equation}
\label{eq:geoNbh}
\Theta_i = \{j | r_{i, j} \le L, r_{i, j} \in R \wedge i \ne j\}.
\end{equation}

\end{definition}

Intuitively, if two grids are of geographical proximity to each other, then there are more chances that they share the same functionality (e.g., two adjacent grids both cover a residential area, where people tend to move out to work in the morning). It should, though, be clear that one grid can not be its own geographical neighbor, since it is meaningless. This neighborhood is useful in clustering semantically similar grids regardless of the behavior of the request stream. When the requests happening between two grids are quite few to be able to provide meaningful information (i.e., the input data is sparse), geographical neighborhood serves as a strong static relationship support. Specifically, the threshold $L$ usually ensures that adjacent grids are geographical neighbors. However, the value can be bigger so that the clustering effect becomes more flexible.

It is essential to notice that the neighborhood sets only provide relationships in low resolution. For example, the number of requests from grid 2 to grid 1 is 26 and that from grid 3 to grid 1 is 105. In this case, grid 2 and grid 3 are both backward neighbors of grid 1, but their strength of neighborhood should absolutely be unequal. The same concern lies in the backward neighborhood and the geographical neighborhood. Hence, it is crucial to add a pre-weighting factor for each neighborhood strength calculation. These factors $a_j^{i, t}$, $b_j^{i, t}$ and $c_j^i$ for forward neighborhood, backward neighborhood and geographical neighborhood correspondingly, are calculated as follows \cite{10.1145/3292500.3330877,Wang2021PassengerMP}:

\begin{equation}
\label{eq:spa_preweights}
\begin{split}
&a_j^{i, t} = \frac{g_{i, j} + \epsilon}{\sum_{k \in \Psi_i^t}{(g_{i, k} + \epsilon)}}, g_{i, j} \in G_t \wedge g_{i, k} \in G_t, \\
&b_j^{i, t} = \frac{g_{j, i} + \epsilon}{\sum_{k \in \Phi_i^t}{(g_{k, i} + \epsilon)}}, g_{j, i} \in G_t \wedge g_{k, i} \in G_t, \\
&c_j^i = \frac{\frac{1}{r_{i, j}}}{\sum_{k \in \Theta_i}{\frac{1}{r_{i, k}}}}, r_{i, j} \in R \wedge r_{i, k} \in R,
\end{split}
\end{equation}
where $\epsilon$ is an extremely small value merely to avoid division by 0.

With the pre-weighting factors specified, the attention weights $\psi_{i, j}^t$, $\phi_{i, j}^t$ and $\theta_{i, j}^t$ are calculated using softmax functions \cite{Wang2021PassengerMP}:
\begin{equation}
\label{eq:spa_att_weights}
\begin{split}
&\psi_{i, j}^t = \frac{\exp(\text{AttentionNet}(\mathrm{v}_i^t, a_j^{i, t}\mathrm{v}_j^t))}{\sum_{k \in \Psi_i^t}{\exp(\text{AttentionNet}(\mathrm{v}_i^t, a_k^{i, t}\mathrm{v}_k^t))}}, \\
&\phi_{i, j}^t = \frac{\exp(\text{AttentionNet}(\mathrm{v}_i^t, b_j^{i, t}\mathrm{v}_j^t))}{\sum_{k \in \Phi_i^t}{\exp(\text{AttentionNet}(\mathrm{v}_i^t, b_k^{i, t}\mathrm{v}_k^t))}}, \\
&\theta_{i, j}^t = \frac{\exp(\text{AttentionNet}(\mathrm{v}_i^t, c_j^i \mathrm{v}_j^t))}{\sum_{k \in \Theta_i}{\exp(\text{AttentionNet}(\mathrm{v}_i^t, c_k^i \mathrm{v}_k^t))}}.
\end{split}
\end{equation}
The attention mechanism is applied using the $\text{AttentionNet}$ function defined as follows \cite{Wang2021PassengerMP}:
\begin{equation}
\label{eq:spa_attnet}
\text{AttentionNet}(\mathrm{v}_i, \mathrm{v}_j) = \text{FC}_a^\mu{(W_a \mathrm{v}_i \oplus W_a \mathrm{v}_j)},
\end{equation}
where $\text{FC}_a^\mu$ denotes a fully-connected layer with the activation function $\mu$ as LeakyReLU and $W_a \in \mathbb{R}^{d_e \times d_f}$ denotes a shared learnable weight matrix to project the feature vectors into the embedding space with dimension $d_e$. The fully connected layer performs a weighted sum of the features from two vectors and provides a scalar as output. It defines a unique perspective to examine the relationship between two grids.

Finally, the weighted features of the neighbors are aggregated together. For grid $i$, the spatial embedding vector $m_i^t$ at time slot $t$ is constructed by concatenating the outputs from the three neighborhoods and the features of grid $i$ together \cite{Wang2021PassengerMP}:
\begin{equation}
\label{eq:spa_embed}
\begin{split}
m_i^t = W_s \mathrm{v}_i^t \oplus \sum_{j \in \Psi_i^t}{\psi_{i, j}^t W_s \mathrm{v}_j^t} \oplus \sum_{j \in \Phi_i^t}{\phi_{i, j}^t W_s \mathrm{v}_j^t} \oplus \\
\sum_{j \in \Theta_i}{\theta_{i, j}^t W_s \mathrm{v}_j^t},
\end{split}
\end{equation}
where $W_s \in \mathbb{R}^{d_e \times d_f}$ denotes a shared learnable weight matrix to project the feature vectors into the embedding space.
%--------------------------------------------------------
Gallat \cite{Wang2021PassengerMP} applies the spatial feature extraction procedure described above. However, the design does not utilize multi-head attention and head gates introduced in GAT \cite{velickovic2018graph} and GaAN \cite{Zhang2018GaANGA}. Essentially, the three neighborhoods generate three graph structures, specifying whether the corresponding relationship exists between two grids. On the other hand, each neighborhood graph generates $K$ attention heads, thus constructing $K$ different perspectives (mainly indicated by the $\text{AttentionNet}$) to examine \textit{how} two grids are related. As an example, suppose grid $j$ is a forward neighborhood of grid $i$, the request flow might be mainly due to a gathering event at grid $j$, or the regular commuting on Friday night. The weights will be applied to the feature dimensions differently for different perspectives (i.e., attention heads). Furthermore, suppose commuting is more common than the gathering event, the attention head considering the commuting case should thus be more important than that considering the gathering event. As a result, the attention gates are utilized.

By extending the spatial attention layer from \cite{Wang2021PassengerMP}, \textit{BGARN} further upgrades Equation \ref{eq:spa_embed} to the following equation:
\begin{equation}
\label{eq:spa_embed_upgrade}
\begin{split}
m_i^t = W_s \mathrm{v}_i^t \  \oplus \parallel_{k=1}^K(W_s \mathrm{v}_i^t + \omega_{i, \Psi_i^t}^k \sum_{j \in \Psi_i^t}{\psi_{i, j}^t W_s \mathrm{v}_j^t}) \  \oplus \\
\parallel_{k=1}^K(W_s \mathrm{v}_i^t + \omega_{i, \Phi_i^t}^k \sum_{j \in \Phi_i^t}{\phi_{i, j}^t W_s \mathrm{v}_j^t}) \  \oplus \\
\parallel_{k=1}^K(W_s \mathrm{v}_i^t + \omega_{i, \Theta_i}^k \sum_{j \in \Theta_i}{\theta_{i, j}^t W_s \mathrm{v}_j^t}),
\end{split}
\end{equation}
where $\omega_{i, \Psi_i^t}^k, \omega_{i, \Phi_i^t}^k, \omega_{i, \Theta_i}^k$ denote the gates for the $k$th head, capturing features of affinity from grid $i$ to its forward neighbors $\Psi_i^t$, backward neighbors $\Phi_i^t$ as well as geographical neighbors $\Theta_i$, correspondingly. $\parallel$ here denotes an aggregation function that can be either an average operation or a sequential concatenation (average is used by default as it consumes less space). BGARN also adds one residual block for each attention output to avoid gradient vanishing when training such deep neural network.

The gates are calculated as follow:
\begin{equation}
\label{eq:spa_gates}
\begin{split}
&\omega_{i, \Psi_i^t} = \text{FC}_{g, \Psi_i^t}^\sigma(\mathrm{v}_i^t \oplus \max_{j \in \Psi_i^t}(W_{g, \Psi_i^t} a_j^{i, t}\mathrm{v}_j^t) \oplus \frac{\sum_{j \in \Psi_i^t}{a_j^{i, t}\mathrm{v_j^t}}}{|\Psi_i^t|}), \\
&\omega_{i, \Phi_i^t} = \text{FC}_{g, \Phi_i^t}^\sigma(\mathrm{v}_i^t \oplus \max_{j \in \Phi_i^t}(W_{g, \Phi_i^t} b_j^{i, t}\mathrm{v}_j^t) \oplus \frac{\sum_{j \in \Phi_i^t}{b_j^{i, t}\mathrm{v_j^t}}}{|\Phi_i^t|}), \\
&\omega_{i, \Theta_i} = \text{FC}_{g, \Theta_i}^\sigma(\mathrm{v}_i^t \oplus \max_{j \in \Theta_i}(W_{g, \Theta_i} c_j^i\mathrm{v}_j^t) \oplus \frac{\sum_{j \in \Theta_i}{c_j^i\mathrm{v_j^t}}}{|\Theta_i|}),
\end{split}
\end{equation}
where $\text{FC}_{g, \Psi_i^t}^\sigma, \text{FC}_{g, \Phi_i^t}^\sigma, \text{FC}_{g, \Theta_i}^\sigma$ represents fully-connected layers with Sigmoid (to generate values between 0 and 1 as gates) as the activation function. They are responsible for mapping the processed vectors into the head space. $W_{g, \Psi_i^t}, W_{g, \Phi_i^t}, W_{g, \Theta_i} \in \mathbb{R}^{d_e \times d_f}$ are three learnable weight matrices to project the pre-weighted feature vectors into the embedding space with dimension $d_e$. $max(\{\mathrm{v}_1, \dots, \mathrm{v}_n\})$ produces the element-wise maximum of each value in the embedded vector. The gating function considers the importance of heads with two perspectives: max pooling and average pooling. After applying the gates, the heads are aggregated to form a further embedded vector of $\mathbb{R}^{d_e}$ (if the average scheme is applied). As a result, the final spatial embedding vector $m_i^t$ is of $\mathbb{R}^{4d_e}$. By stacking these vectors vertically, the spatial embedding matrix $M_t = [m_1^t, \dots, m_n^t]^T \in \mathbb{R}^{n \times 4d_e}$ is retrieved. For an OD graph sequence $\{G_t\}_{t=1}^T$, the Spatial Attention Later will obtain the corresponding embedding matrices $\{M_t\}_{t=1}^T$.

\begin{figure}
    \centerline{\includegraphics[width=3.5in]{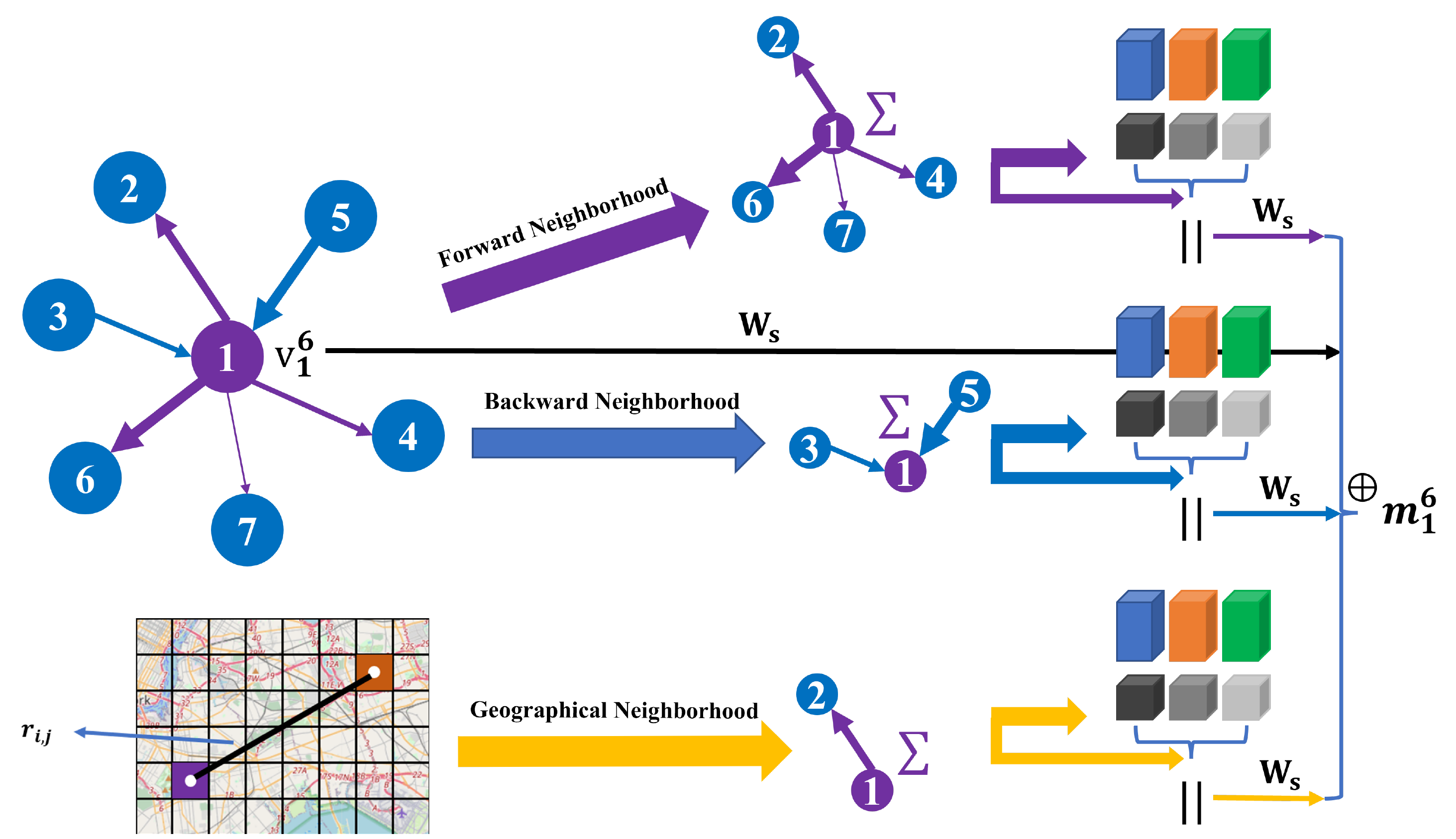}}
    \caption{\textbf{Spatial Attention Layer}: In the example, the spatial embedding vector $m_1^6$ of grid $1$ is calculated. First, three neighborhoods are extracted respectively. Then, each generates $K$ (here $K=3$) attention heads. Finally, all aggregated heads are concatenated again along with the initial projected feature vector $\mathrm{v}_1^6$. $\sum$ specifies the attention aggregation operation.}
    \label{fig:spatial}
\end{figure}

Figure \ref{fig:spatial} summarizes the key operations in the spatial attention layer.

\subsection{Temporal Recurrent Layer}
\label{sec:tempreclayer}

As suggested in \cite{10.5555/3298239.3298479}, there are two main aspects of temporal features: tendency and periodicity, which introduce short-term as well as long-term time dependencies respectively.

\begin{definition}[Tendency]
\label{def:tendency}

The request flow pattern is easily affected by those from the past few time slots due to the continuity of time. The set of spatial embedding matrices to be considered to this concern is as follows \cite{10.5555/3298239.3298479,Wang2021PassengerMP}:
\begin{equation}
\label{eq:St}
S_t = \{M_t | t = T + 1 - p, p \in [1, P]\},
\end{equation}
where $P$ specifies the total number of historical records considered.

\end{definition}

\begin{definition}[Periodicity]
\label{def:periodicity}

The request flow pattern appears to be similar to those from the same time slot in the past few days due to the daily mobility behavior. The set of spatial embedding matrices to be considered to this concern is as follows \cite{10.5555/3298239.3298479,Wang2021PassengerMP}:
\begin{equation}
\label{eq:Sp}
S_p = \{M_t | t = T + 1 - lp, p \in [1, P]\},
\end{equation}
where $l$ specifies the number of time slots per day and $P$ specifies the total number of historical records considered.

\end{definition}

Note that RSODP intends to extract the temporal features for predicting the requests in time slot $T+1$, so the subtractions in Equation \ref{eq:St} and \ref{eq:Sp} are based on $T+1$. Besides, it is required that $P \le \lfloor \frac{T}{l} \rfloor$.

Intuitively, as an example of tendency, if there is a considerable number of people moving from grid $1$ (residential area) to grid $2$ (workplace) to work at 8:00 a.m., the request flow pattern might persist in the short future, meaning that there are probably many people moving from grid $1$ to grid $2$ to work at 9:00 a.m. as well; as for periodicity, it is common that people leave home to work nearly at the same time of each workday. In light of this, it can be hypothesized that the request flow pattern tomorrow at the same time is rather likely to be similar to that of today.

To have a miscellaneous view of both tendency and periodicity, BGARN also considers the prior and posterior time slot of the periodic time sequence $S_p$, as suggested in \cite{Wang2021PassengerMP}:
\begin{equation}
\label{eq:Stp}
\begin{split}
&S_{tp^-} = \{M_t | t = T - lp, p \in [1, P]\}, \\
&S_{tp^+} = \{M_t | t = T + 2 - lp, p \in [1, P]\}.
\end{split}
\end{equation}

Gallat \cite{Wang2021PassengerMP} uses Scaled Dot-Product Attention in their Temporal Attention Layer design. By multiplying query features with spatial embeddings (keys), followed by a row-wise softmax function, the dispatching pattern of requests from one grid to the others can be examined. Nevertheless, such technique serves more as a supplemental spatial extraction, since it operates on the grids instead of the timeline. The actual temporal feature extraction in Gallat is a simple summation. By comparison, BGARN feeds the spatial embeddings sequentially into a recurrent module such as LSTM, then average the results over all temporal dimensions.

Basically, a time sequence $S_x \in \{S_t, S_p, S_{tp^-}, S_{tp^+}\}$ is forwarded to a recurrent network (e.g., LSTM) to generate the temporal embedding matrix $M_{S_x}$:
\begin{equation}
\label{eq:Msx}
\begin{split}
M_{S_x} = \text{LSTM}(S_x),\ S_x \in \{S_t, S_p, S_{tp^-}, S_{tp^+}\}.
\end{split}
\end{equation}

From Equation \ref{eq:Msx}, four feature embeddings are retrieved from different temporal dimensions. By aggregating them (average by default), the Temporal Recurrent Layer retrieves the spatial-temporal feature embedding matrix $M'_T \in \mathbb{R}^{n \times 4d_e}$ as follow:
\begin{equation}
\label{eq:MTprime}
\begin{split}
M'_T = \text{BN}(\parallel_{M_{S_x} \in \{M_{S_t}, M_{S_p}, M_{S_{tp^-}}, M_{S_{tp^+}}\}}{M_{S_x}}),
\end{split}
\end{equation}
where $\text{BN}$ specifies the Batch Normalization operation.

\begin{figure}
    \centerline{\includegraphics[width=3.5in]{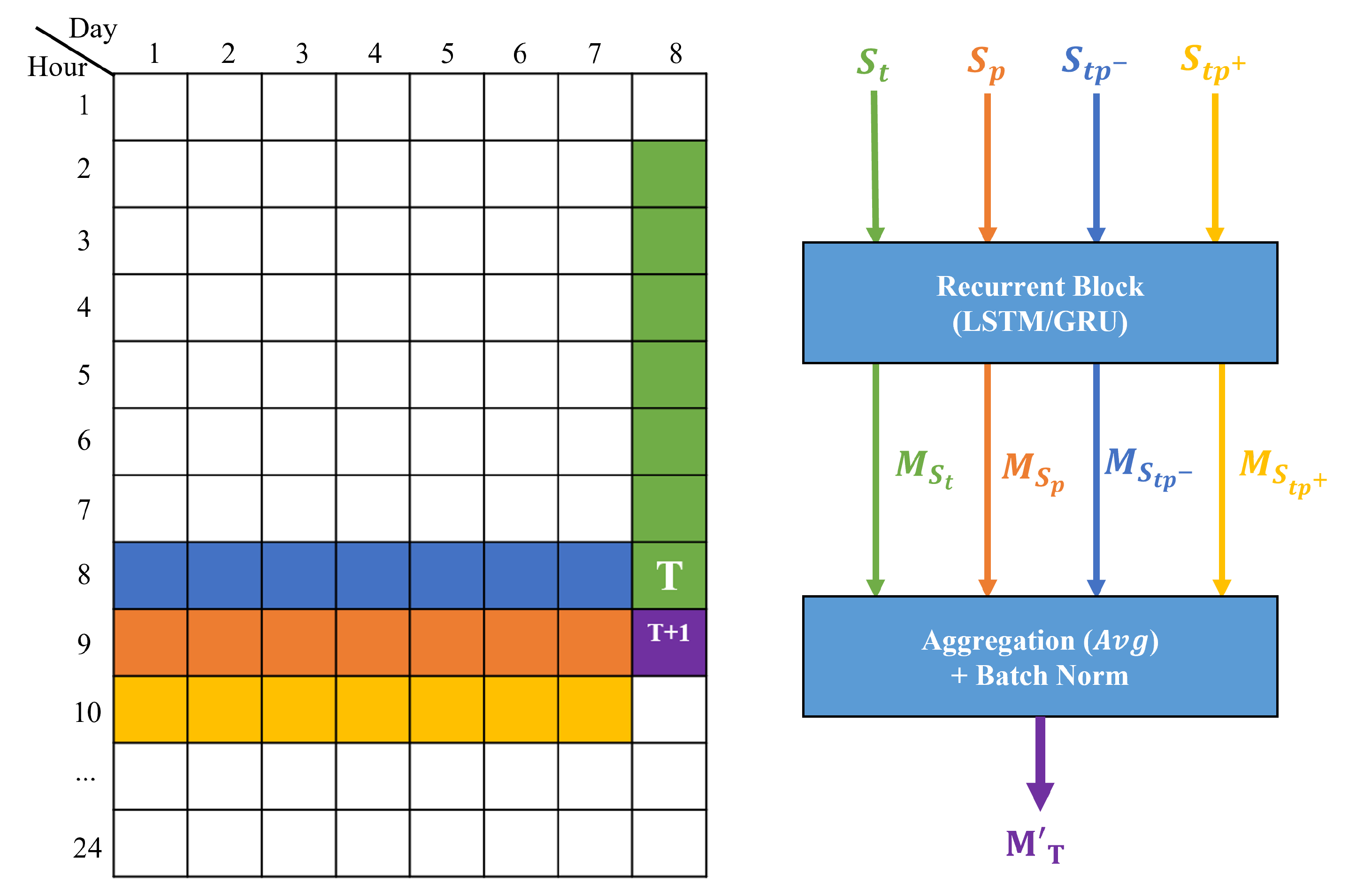}}
    \caption{Temporal Recurrent Layer. Four time series regarding the tendency, periodicity of time are extracted. They are passed into the recurrent modules and aggregated to output the spatial-temporal embedding matrix $M'_T$.}
    \label{fig:temporal}
\end{figure}

Figure \ref{fig:temporal} summarizes the key operations in the temporal recurrent layer.

\subsection{Transferring Layer}

The Transferring Layer utilizes the spatial-temporal embedding matrix $M'_T$ for two prediction tasks: the demand task which predicts the number of outgoing requests $\hat{d}_{T+1}$ starting from each grid, and the OD task which predicts the OD graph $\hat{G}_{T+1}$ in the next time slot. Compared to the OD task, the demand task predicts only the origin of the request flow, thus largely reduces the complexity (from $n^2$ predictions to $n$). Therefore, it will be helpful to set the demand task as a sub-task.

BGARN utilises simple baseline models, like HA (Historical Average) and AR (Auto-Regressive) described in section \ref{Baselines, Other Models}, to provide a rough estimate upon which the deep learning model can improve its predictions. By performing an operation (termed in this paper as \textit{tuning}) which combines the baseline results (linearity) and the prediction results inferred from $M'_T$ (non-linearity), the Transferring Layer obtains the final prediction outputs. The equations for the tasks are as follows:
\begin{equation}
\label{eq:tasks}
\begin{split}
&\hat{d}_{T+1} = \text{Aggr}(\text{FC}_{d}^{l}(M'_T), \hat{d}_{T+1}^{ref}), \\
&\hat{g}_{i, j} = \text{Aggr}(\text{AttentionNet}({m'}_i^T, {m'}_j^T), \hat{g}_{i, j}^{ref}), \\
&\text{Aggr}_{[\text{sum}]}(a, b) = a + b, \\
&\text{Aggr}_{[\text{wsum}]}(a, b) = w a + (1-w) b,\ w \in [0, 1] \\
&\text{Aggr}_{[\text{mult}]}(a, b) = a \times b,
\end{split}
\end{equation}
where $\hat{d}_{T+1}^{ref}$ and $\hat{g}_{i, j}^{ref}$ are baseline outputs and $\text{Aggr}(a, b)$ specifies the tuning approach which can be sum, weighted sum or multiplication (multiplication by default). Basically, for weighted sum, both deep learning results and baseline results represent the request flow. The deep learning results for sum and multiplication, however, have a different meaning. In these two cases, they serve as a tuning factor of the baseline outputs, which might be more appropriate since normalization normally scales the intermediate values to around 1.0.

\section{Experimental Evaluation and Results}
\label{sec:experiment}

In this section, BGARN is evaluated to examine whether it performs the predictions on demands and request graphs effectively as expected.

\subsection{Dataset}

\setlength{\arrayrulewidth}{1pt}
\setlength{\tabcolsep}{12pt}
\renewcommand{\arraystretch}{1.35}
\begin{table}
    \caption{Summary for the New York Yellow Taxi Trip data}
    \label{tab:dataset}
    \centering
    \begin{tabular}{| c || c |}
        \hline
        Dataset & New York Yellow Taxi Trip (2016) \\
        \hhline{| = || = |}
        Time span & 3 months (2184 hours) \\
        \hline
        Total area & $47.31 \times 47.88\ \text{km}^2$ \\
        \hline
        Partition scheme & rectangle \\
        \hline
        Grid distribution & $19 \times 19$ \\
        \hline
        Grid granularity & $2.49 \times 2.52\ \text{km}^2$ \\
        \hline
        Time slot granularity & 1 hour \\
        \hline
    \end{tabular}
\end{table}

Table \ref{tab:dataset} summarizes the dataset. The New York Yellow Taxi Trip\footnote{Data URL: \url{https://www.kaggle.com/vishnurapps/newyork-taxi-demand}.} data is selected to conduct the experiments. Basically, a portion of the data from Jan 1st, 2016 to Mar 31st, 2016 (3 months of time span) is collected and preprocessed. For a vehicle with speed $30$ km/h, it takes around 5 minutes to travel through a grid, which is often considered to be an acceptable waiting time for the passengers. In light of this, New York City is partitioned into 361 grids, each with size of $2.49 \times 2.52\ \text{km}^2$. the requests are split by 1-hour granularity since the human mobility pattern is often summarized in hours. For example, it is common to mention the phrase "rush hours", which specifies the hours in a day when traffic is the heaviest. Besides, 1 hour of time should be enough for the prediction model to perform one-round gradient descent and then for the optimizer to dispatch vehicle-request assignments.

\subsection{Baselines, Other Models \& Variants}
\label{Baselines, Other Models}

The metrics evaluation results of BGARN are compared with the following baseline models:
\begin{itemize}
    \item \textbf{HA}: Historical Average is the very baseline method which computes the average of the historical demands from the previous time slots. Three versions of HA are tested using different temporal features settings, with HA$^+$ using all four as described in section \hyperref[sec:tempreclayer]{1.4.3}, HAt using only tendency and HAp using only periodicity.
    
    \item \textbf{AR}: AR (Auto-regressive) model. This paper uses a simple feed-forward network which calculates a weighted sum of the historical data.
\end{itemize}

\noindent In addition BGARN is contrasted with four existing state of the art models LSTNet \cite{lai2018modeling},  GCRN\cite{Seo2018StructuredSM}, GEML \cite{10.1145/3292500.3330877} and Gallat \cite{Wang2021PassengerMP}.

%\begin{itemize}
 %   \item \uses a short-term convolution and a GRU to aggregate temporal features, along with an attention block to leverage the historical records along tendency.
    
 %   \item \textbf uses graph convolution to extract spatial features and LSTM to extract temporal features along tendency.
    
 %   \item \textbf{ uses pre-weighted GAT to extract spatial features and periodic-skip LSTM to extract temporal features along periodicity.
    
  %  \item \textbf uses pre-weighted GAT to extract spatial features and Scaled Dot-Product Attention to extract temporal features along both tendency and periodicity. 
%\end{itemize}

The default aggregation scheme of BGARN is set as average in both spatial and temporal layer (since it consumes less memory). The default aggregation approach with the baseline results is multiplication with baseline results. Further, This paper specifies four variants for BGARN:
\begin{itemize}
    \item \textbf{BGARN-NoTune}: inherits the design of the transferring layer in Gallat, meaning there is no tuning with the baseline results.
    
    \item \textbf{BGARN-Concat}: uses concatenation as the aggregation scheme in both spatial and temporal layer.
    
    \item \textbf{BGARN-WSum}: uses another tuning approach - weighted sum which adds the results from the attention layers and those from the baseline algorithm together with scaling weights specified.
    
    \item \textbf{BGARN-Shift}: uses another tuning approach - shifting (sum) which adds the results from the attention layers and those from the baseline algorithm together.
\end{itemize}

\subsection{Parameter Settings}

Smooth L1 Loss is used to calculate an overall loss as follow, as suggested in \cite{Wang2021PassengerMP}:
\begin{equation}
\label{eq:loss}
\begin{split}
&\mathcal{L} = \eta_d \mathcal{L}_d + \eta_o \mathcal{L}_o, \\
&\mathcal{L}_d = \text{SmoothL1Loss}(\hat{d}_{T+1}, d_{T+1}), \\
&\mathcal{L}_o = \text{SmoothL1Loss}(\hat{G}_{T+1}, G_{T+1}),
\end{split}
\end{equation}
where $\eta_d$ and $\eta_o$ are two hyper-parameters to balance the importance of each task. The reason to use Smooth L1 Loss rather than Mean Square Error Loss is that it is more robust to outliers than L2 Loss so that the gradients do not change drastically when encountering abnormal data input, as explained in \cite{7410526}.

For metrics evaluation, this paper adopts the three classic functions - RMSE (Root Mean Square Error), MAPE (Mean Absolute Percentage Error) and MAE (Mean Absolute Error), which are widely used in regression tasks. The formulas are as follows:
\begin{equation}
\label{eq:metrics}
\begin{split}
&\text{RMSE}(y, \hat{y}) = \sqrt{\frac{1}{z}\sum_{i=1}^{z}(y - \hat{y})^2}, \\
&\text{MAPE}(y, \hat{y}) = \frac{1}{z}\sum_{i=1}^{z}|\frac{y - \hat{y}}{y + 1}|, \\
&\text{MAE}(y, \hat{y}) = \frac{1}{z}\sum_{i=1}^{z}|y - \hat{y}|,
\end{split}
\end{equation}
where $z$ specifies the number of batches. $\hat{y}$ and $y$ specify the predicted results and ground true values respectively.

For the experiments, Adam has been used as the optimizer and all the models are trained (implemented \footnote{Implementation: \url{https://github.com/WingsUpete/RSODP}.} using PyTorch \cite{NEURIPS2019_9015} and the Deep Graph Library \cite{wang2019dgl}) on Tesla P100 PCIe. The settings of training epochs, batch size, gradient clipping norm, hidden dimension, number of attention heads, number of historical records $P$, task importances $\eta_d$ and $\eta_o$ are respectively 200, 32, 10.0, 16, 3, 7, 0.8, 0.2.

\subsection{Results}

\setlength{\tabcolsep}{3pt}
\renewcommand{\arraystretch}{1.35}
\begin{table}
    \caption{TTpS (Training Time per Sample) results}
    \label{tab:resultsTTpS}
    \centering
    \begin{tabular}{| c | c | c | c | c | c |}
        \hline
        Model & LSTNet & GCRN & GEML & Gallat & \underline{BGARN} \\
        \hline
        TTpS & 0.0965 sec & 0.0497 sec & 0.0387 sec & 0.1579 sec & 0.2631 sec \\
        \hline
    \end{tabular}
\end{table}

\begin{table}
\caption{RMSE results for the models}
\label{tab:resultsRMSE}
\centering
\begin{tabular}{| c | c || c | c | c |}
\hline
\multirow{2}{*}{Task} & \multirow{2}{*}{Model} & \multicolumn{3}{c|}{New York Yellow Taxi Trip (2016)} \\
\cline{3-5}
& & RMSE-0 & RMSE-3 & RMSE-5 \\
\hline
\multirow{13}{*}{Demand} & HA$^+$ & 218.7742 & 289.7921 & 309.9055 \\
& HAt & 422.4178 & 559.4045 & 598.1556 \\
& HAp & 203.6360 & 269.7786 & 288.5173 \\
& AR & 124.7427 & 165.2147 & 176.6445 \\
& LSTNet & 124.3148 & 164.6435 & 176.0304 \\
& GCRN & 161.4843 & 213.9025 & 228.7444 \\
& GEML & 142.5458 & 188.7993 & 201.8993 \\
& Gallat & 897.7490 & 1189.3735 & 1272.0073 \\
\cline{2-5}
& \underline{BGARN} & \textbf{86.6160} & \textbf{114.7069} & \textbf{122.6550} \\
\cline{2-5}
& BGARN-NoTune & 93.3746 & 123.6710 & 132.2483 \\
& BGARN-Concat & 91.4094 & 121.0523 & 129.4348 \\
& BGARN-WSum & 100.9402 & 133.7110 & 142.9936 \\
& BGARN-Shift & 92.8987 & 123.0300 & 131.5582 \\
\hline
\multirow{13}{*}{OD} & HA$^+$ & 28.3008 & 48.5195 & 55.5404 \\
& HAt & 53.4073 & 91.5448 & 104.7805 \\
& HAp & 26.6731 & 45.7376 & 52.3616 \\
& AR & 25.4056 & 43.5492 & 49.8539 \\
& LSTNet & 25.2567 & 43.2953 & 49.5642 \\
& GCRN & 105.9160 & 181.7916 & 208.2482 \\
& GEML & 46.4638 & 79.5246 & 91.0136 \\
& Gallat & 105.9155 & 181.7907 & 208.2473 \\
\cline{2-5}
& \underline{BGARN} & \textbf{18.6892} & \textbf{32.0242} & \textbf{36.6500} \\
\cline{2-5}
& BGARN-NoTune & 105.7277 & 181.4710 & 207.8826 \\
& BGARN-Concat & \textbf{17.0450} & \textbf{29.1927} & \textbf{33.3987} \\
& BGARN-WSum & 86.6196 & 148.6731 & 170.3114 \\
& BGARN-Shift & 28.2729 & 48.4722 & 55.4867 \\
\hline
\end{tabular}
\end{table}

\begin{table}
\caption{MAPE results for the models}
\label{tab:resultsMAPE}
\centering
\begin{tabular}{| c | c || c | c | c |}
\hline
\multirow{2}{*}{Task} & \multirow{2}{*}{Model} & \multicolumn{3}{c|}{New York Yellow Taxi Trip (2016)} \\
\cline{3-5}
& & MAPE-0 & MAPE-3 & MAPE-5 \\
\hline
\multirow{13}{*}{Demand} & HA$^+$ & 0.5244 & 0.4837 & 0.4601 \\
& HAt & 0.9104 & 0.8841 & 0.8498 \\
& HAp & 0.4193 & 0.3869 & 0.3691 \\
& AR & 0.5597 & 0.4111 & 0.3641 \\
& LSTNet & 0.5702 & 0.4175 & 0.3690 \\
& GCRN & 0.4327 & 0.3706 & 0.3467 \\
& GEML & 1.1417 & 0.5944 & 0.5088 \\
& Gallat & 0.6344 & 0.8944 & 0.9305 \\
\cline{2-5}
& \underline{BGARN} & \textbf{0.3983} & \textbf{0.3242} & \textbf{0.2939} \\
\cline{2-5}
& BGARN-NoTune & 0.5364 & 0.3379 & 0.2994 \\
& BGARN-Concat & 0.4079 & 0.3290 & 0.2954 \\
& BGARN-WSum & 0.4616 & 0.3163 & 0.2918 \\
& BGARN-Shift & 0.4833 & 0.3534 & 0.3073 \\
\hline
\multirow{13}{*}{OD} & HA$^+$ & 0.4213 & 0.4099 & 0.3966 \\
& HAt & 0.6176 & 0.7066 & 0.7069 \\
& HAp & 0.4000 & 0.3685 & 0.3497 \\
& AR & 0.4403 & 0.3798 & 0.3528 \\
& LSTNet & 0.4363 & 0.3752 & 0.3481 \\
& GCRN & 0.6807 & 0.9082 & 0.9377 \\
& GEML & 0.8023 & 0.7917 & 0.7608 \\
& Gallat & 0.6799 & 0.9079 & 0.9375 \\
\cline{2-5}
& \underline{BGARN} & \textbf{0.3992} & \textbf{0.3672} & \textbf{0.3439} \\
\cline{2-5}
& BGARN-NoTune & 0.6182 & 0.8793 & 0.9159 \\
& BGARN-Concat & 0.4128 & 0.3759 & 0.3487 \\
& BGARN-WSum & 0.5097 & 0.6955 & 0.7227 \\
& BGARN-Shift & 0.4194 & 0.4081 & 0.3950 \\
\hline
\end{tabular}
\end{table}

\begin{table}
\caption{MAE results for the models}
\label{tab:resultsMAE}
\centering
\begin{tabular}{| c | c || c | c | c |}
\hline
\multirow{2}{*}{Task} & \multirow{2}{*}{Model} & \multicolumn{3}{c|}{New York Yellow Taxi Trip (2016)} \\
\cline{3-5}
& & MAE-0 & MAE-3 & MAE-5 \\
\hline
\multirow{13}{*}{Demand} & HA$^+$ & 61.3402 & 106.5441 & 121.3612 \\
& HAt & 115.1495 & 200.2938 & 228.2327 \\
& HAp & 53.0514 & 92.2172 & 105.0758 \\
& AR & 37.4329 & 64.2452 & 72.9105 \\
& LSTNet & 37.4633 & 64.2670 & 72.9223 \\
& GCRN & 42.5027 & 73.6084 & 83.7732 \\
& GEML & 44.3198 & 74.3691 & 84.1335 \\
& Gallat & 263.2721 & 461.4646 & 527.3090 \\
\cline{2-5}
& \underline{BGARN} & \textbf{26.7627} & \textbf{46.0133} & \textbf{52.2123} \\
\cline{2-5}
& BGARN-NoTune & 28.9615 & 49.3438 & 55.9659 \\
& BGARN-Concat & 27.6920 & 47.6142 & 54.0222 \\
& BGARN-WSum & 31.8131 & 54.6169 & 62.0902 \\
& BGARN-Shift & 27.6424 & 47.2541 & 53.5217 \\
\hline
\multirow{13}{*}{OD} & HA$^+$ & 5.8884 & 15.2942 & 19.3079 \\
& HAt & 10.5099 & 28.1678 & 35.7743 \\
& HAp & 5.3692 & 13.8239 & 17.4191 \\
& AR & 5.4992 & 13.9453 & 17.5141 \\
& LSTNet & 5.4623 & 13.8542 & 17.4021 \\
& GCRN & 20.5310 & 57.6906 & 74.3360 \\
& GEML & 11.6991 & 30.5545 & 38.5964 \\
& Gallat & 20.5283 & 57.6879 & 74.3333 \\
\cline{2-5}
& \underline{BGARN} & \textbf{4.7045} & \textbf{11.8547} & \textbf{14.8125} \\
\cline{2-5}
& BGARN-NoTune & 20.2589 & 57.2617 & 73.8595 \\
& BGARN-Concat & \textbf{4.5955} & \textbf{11.4526} & \textbf{14.2488} \\
& BGARN-WSum & 16.3076 & 45.9895 & 59.3267 \\
& BGARN-Shift & 5.8736 & 15.2601 & 19.2670 \\
\hline
\end{tabular}
\end{table}

As mentioned in \cite{Wang2021PassengerMP}, the model should focus more on the regions with higher number of requests generated. Therefore, three thresholds are specified - 0, 3, 5, to filter outputs below these thresholds and only calculate the metrics on the filtered outputs (e.g., MAE-5 specifies the Mean Absolute Error with threshold as 5).

The results are shown in table \ref{tab:resultsTTpS}, \ref{tab:resultsRMSE}, \ref{tab:resultsMAPE} and \ref{tab:resultsMAE}. Derived from table \ref{tab:resultsTTpS}, it appears to be quite time-efficient to pass one sample (one-hour request data) through the model. Furthermore, the training time of BGARN does not increase linearly with regard to the number of attention heads (Gallat can be considered as using only one attention head).

From table \ref{tab:resultsRMSE}, \ref{tab:resultsMAPE} and \ref{tab:resultsMAE}, it can be inferred that the RMSE, MAPE and MAE results of BGARN are best among those of all models for the Demand task and the OD task. In general, the metrics specifying errors for the OD task are significantly lower than those for the Demand task. This is because the request graph is rather sparse and the error value for each slot in the results is diluted.

The results of the baseline models are surprisingly good. This can be on account of the fact that the utilized dataset possesses a strong periodic feature, as indicated by the results from HAp. Simply averaging the weighted historical values (AR) seems to already provide a strong prediction output. Nevertheless, BGARN still manages to provide more accurate weights while some other models fail to. One of the most important reasons is that BGARN sets the outputs from baseline models as a basis and then improve them with refined feature extractions.

By column-wise comparison, MAPE values generally drop with increasing thresholds. This might be due to the larger denominators when calculating percentages with more requests. On the other hand, RMSE and MAE generally increase, which indicates the complexity of predicting the request flow at locations with heavy traffic.

\subsection{Comparative Analysis}

Generally, BGARN and its variants appear to perform better than those of Gallat, which indicates that the GaAN design in the spatial layer of BGARN is able to capture better feature representations than GAT used in Gallat. It is, in the meantime, interesting to note that Gallat and BGARN-NoTune, which are both designed without tuning in the transferring layer, perform far worse than the others. As a result, it seems to be rather helpful to combine baseline results with the model.

It is also noticeable that LSTNet performs better than GCRN, GEML and Gallat. This might be because LSTNet processes the request matrices ($d$ and $G$) directly, while GEML and Gallat do not. Instead, GEML and Gallat uses the request matrices to generate features and different graph views. Although GCRN utilizes the request matrices in the spatial feature extraction module, it considers the historical records along tendency, which leads to a worse set of results with a highly periodic dataset. GEML, on the other hand, uses the historical records along periodicity, thus retrieve better results than GCRN. It is worth noticing that GEML, though uses a combined request view compared to Gallat, still manages to outperform Gallat and BGARN-NoTune, indicating that a simple GRU or LSTM might be better than using Scaled Dot-Product Attention.

\subsection{Analysis of Variants}

BGARN-NoTune performs far worse than BGARN, indicating that the baseline results are rather helpful in providing more accurate results. BGARN-WSum and BGARN-Shift, however, perform even worse than BGARN-NoTune in the Demand task. The reason might be that the outputs of the deep learning model before tuning are actually small (around 1.0). In this case, tuning by scaling should be the best scheme among all. In the Demand task, the values are much larger so that the shifting scheme performs much worse than it does in the OD task. This also helps explain why Gallat (with no tuning) does not perform well, as it uses a sigmoid activation in the Transferring Layer which stagnates the training process when outputting 1.

Finally, BGARN-Concat, which uses concatenation as the aggregation scheme in the spatial and temporal layer, performs slightly worse than the average version on Demand Task, while on OD Task it performs better. The results again show that OD prediction is a far more complicated task compared to predicting only the origin. Furthermore, it is found that BGARN-Concat provided more stable results through extensive repeated experiments, which makes sense since it preserves all the features without reduction. Nevertheless, direct concatenation consumes much more space and trains significantly longer, thus it might not be suitable for use in real-word cases.

%--------------------------------------
\section{Conclusions}
\label{sec:conclusion}
%--------------------------------------

This paper has revisited the concept of Ridesharing  and has proposed a new model, BGARN, for addressing the Origin-Destination Prediction for Ridesharing (RSODP) challenge. The novel features of GARN are the utilisation of multi-head gated attention and a tuning approach  which combines linear baseline results with non-linear deep learning results. 

The utilization of multi-head gated attention provides an integrated view of different request flow relationship measurements among grids. This enables the capturing of multiple perspectives as well as their corresponding importance, thus supporting a more holistic analysis of the request flow and delivering more accurate predictions. In addition, the proposed tuning approach significantly enhances the prediction capability of the model. This approach is generic and can be applied to any regression task.

The experimental results obtained using the on the New York Yellow
Taxi Trip dataset confirm that BGARN outperforms all the existing state of the art models in terms of prediction accuracy.

In the future, the model will be further extended by applying hexagon-based grid partitioning in the Preprocessing Module. It will also be tested on larger datasets (such as request streams in Beijing and Shanghai) with more complicated request dynamics.

% Appendix
\section*{Appendix}

Table \ref{tab:notations} lists and explains the notations used in the paper.  

\setlength{\tabcolsep}{2pt}
\renewcommand{\arraystretch}{1.35}
\begin{table}
    \caption{Notations used in the paper}
    \label{tab:notations}
    \centering
    \begin{tabular}{| c | c |}
        \hline
        Symbol & Meaning \\
        \hline
        $\oplus$ & concatenation operation \\
        \hline
        $\parallel$ & aggregation function \\
        \hline
        $t$ & a time slot \\
        \hline
        $t_n$ & \makecell{time endurance of a time slot \\ in hours, usually set as 1} \\
        \hline
        $l$ & number of time slots per day \\
        \hline
        $T$ & total number of time slots of the input sequence \\
        \hline
        $G_t$ & OD graph at time slot $t$ \\
        \hline
        $n$ & number of grids for all $G_t$ \\
        \hline
        $\mathcal{V}_t, \mathcal{E}_t$ & grid and edge set for OD graph $G_t$ \\
        \hline
        $v_i^t$ & grid $i$ at time slot $t$ \\
        \hline
        $e_{i, j}^t$ & \makecell{edge representing an OD flow from grid $i$ \\ to grid $j$ for OD graph $G_t$} \\
        \hline
        $g_{i, j}^t$ & \makecell{the number of requests submitted \\ from grid $i$ to grid $j$ at time slot $t$} \\
        \hline
        $R$ & \makecell{geographical adjacency matrix recording the \\ haversine distances among grids for all $G_t$} \\
        \hline
        $d$ & a request \\
        \hline
        $t_r$ & \makecell{request time (when the request \\ is submitted to the system)} \\
        \hline
        $\text{lat}$, $\text{lng}$ & latitudes and longitudes of the origin/destination \\
        \hline
        $\mathcal{D}$ & request set, aka the input sequence \\
        \hline
        $d_{T+1}$, $\hat{d}_{T+1}$ & \makecell{ground truth and the predicted demand vector \\ storing the number of outgoing requests \\ from each grid} \\
        \hline
        $\hat{G}_{T+1}$ & \makecell{the predicted OD graph} \\
        \hline
        $\mathrm{v}_i^t$ & feature vector for grid $i$ at time slot $t$ \\
        \hline
        $V_t$ & grid feature matrix for all grids at time slot $t$ \\
        \hline
        $\Psi_i^t$, $\Phi_i^t$, $\Theta_i$ & \makecell{set of forward, backward, geographical \\ neighbors of grid $i$ at time slot $t$} \\
        \hline
        $L$ & geographical neighboring threshold \\
        \hline
        $a_j^{i, t}$, $b_j^{i, t}$, $c_j^i$ & \makecell{pre-weights of forward, backward, geographical \\ neighbor $j$ of grid $i$ at time slot $t$} \\
        \hline
        $\epsilon$ & \makecell{an extremely small value to avoid \\ denominator of a fraction being 0} \\
        \hline
        $\psi_{i, j}^t$, $\phi_{i, j}^t$, $\theta_{i, j}^t$ & \makecell{attention weights of forward, backward, \\ geographical neighborhood affinity \\ from grid $i$ to grid $j$ at time slot $t$} \\
        \hline
        $m_i^t$ & spatial embedding vector of grid $i$ at time slot $t$ \\
        \hline
        $M_t$ & spatial embedding matrix of all grids at time slot $t$ \\
        \hline
        $K$ & number of attention heads \\
        \hline
        $\omega_{i, \Psi_i^t}^k$, $\omega_{i, \Phi_i^t}^k$, $\omega_{i, \Theta_i}^k$ & \makecell{gate vector for the $k$th head, capturing features \\ of affinity from grid $i$ to its forward neighbors $\Psi_i^t$, \\ backward neighbors $\Phi_i^t$ as well as geographical \\ neighbors $\Theta_i$} \\
        \hline
        $P$ & number of historical records to be considered \\
        \hline
        $S_t$, $S_p$, $S_{tp^-}$, $S_{tp^+}$ & \makecell{tendency, periodicity and miscellaneous \\ time series of spatial embedding matrices} \\
        \hline
        $M_{S_x}$ & temporal embedding matrix for time series $S_x$ \\
        \hline
        $M'_T$ & spatial-temporal embedding matrix \\
        \hline
        $d_f$, $d_e$ & \makecell{dimension of the feature vector, \\ spatial embedding vector} \\
        \hline
        $\mu$ & LeakyReLU activation function \\
        \hline
        $\sigma$ & Sigmoid activation function \\
        \hline
        $\varrho$ & Row-wise Softmax \\
        \hline
        $W$ & learnable matrices \\
        \hline
    \end{tabular}
\end{table}

% Acknowledgment

% \section*{Acknowledgements}

% This research was supported by: Shenzhen Science and Technology Program,  China (No. GJHZ20210705141807022); Guangdong Province Innovative and Entrepreneurial Team Programme, China (No. 2017ZT07X386); SUSTech Research Institute for Trustworthy Autonomous Systems, China.

%%% Reference %%%
\renewcommand\refname{REFERENCES}
\bibliographystyle{IEEEtran}
\bibliography{tex}

\begin{thebibliography}{10}
\providecommand{\url}[1]{#1}
\csname url@samestyle\endcsname
\providecommand{\newblock}{\relax}
\providecommand{\bibinfo}[2]{#2}
\providecommand{\BIBentrySTDinterwordspacing}{\spaceskip=0pt\relax}
\providecommand{\BIBentryALTinterwordstretchfactor}{4}
\providecommand{\BIBentryALTinterwordspacing}{\spaceskip=\fontdimen2\font plus
\BIBentryALTinterwordstretchfactor\fontdimen3\font minus
  \fontdimen4\font\relax}
\providecommand{\BIBforeignlanguage}[2]{{%
\expandafter\ifx\csname l@#1\endcsname\relax
\typeout{** WARNING: IEEEtran.bst: No hyphenation pattern has been}%
\typeout{** loaded for the language `#1'. Using the pattern for}%
\typeout{** the default language instead.}%
\else
\language=\csname l@#1\endcsname
\fi
#2}}
\providecommand{\BIBdecl}{\relax}
\BIBdecl

\bibitem{Mitropoulos}
\BIBentryALTinterwordspacing
L.~Mitropoulos, A.~Kortsari, and G.~Ayfantopoulou, ``A systematic literature
  review of ride-sharing platforms, user factors and barriers,'' \emph{Eur.
  Transp. Res. Rev.}, vol.~13, no.~61, 2021. [Online]. Available:
  \url{https://doi.org/10.1186/s12544-021-00522-1}
\BIBentrySTDinterwordspacing

\bibitem{2021arXiv210111174J}
W.~Jiang and J.~Luo, ``{Graph Neural Network for Traffic Forecasting: A
  Survey},'' \emph{arXiv e-prints}, p. arXiv:2101.11174, Jan. 2021.

\bibitem{10.14778/3339490.3339493}
\BIBentryALTinterwordspacing
J.~J. Pan, G.~Li, and J.~Hu, ``Ridesharing: Simulator, benchmark, and
  evaluation,'' \emph{Proc. VLDB Endow.}, vol.~12, no.~10, p. 1085–1098, Jun.
  2019. [Online]. Available: \url{https://doi.org/10.14778/3339490.3339493}
\BIBentrySTDinterwordspacing

\bibitem{lai2018modeling}
\BIBentryALTinterwordspacing
G.~Lai, W.~Chang, Y.~Yang, and H.~Liu, ``Modeling long- and short-term temporal
  patterns with deep neural networks,'' \emph{CoRR}, vol. abs/1703.07015, 2017.
  [Online]. Available: \url{http://arxiv.org/abs/1703.07015}
\BIBentrySTDinterwordspacing

\bibitem{Seo2018StructuredSM}
Y.~Seo, M.~Defferrard, P.~Vandergheynst, and X.~Bresson, ``Structured sequence
  modeling with graph convolutional recurrent networks,'' \emph{ArXiv}, vol.
  abs/1612.07659, 2018.

\bibitem{10.1145/3292500.3330877}
\BIBentryALTinterwordspacing
Y.~Wang, H.~Yin, H.~Chen, T.~Wo, J.~Xu, and K.~Zheng, ``Origin-destination
  matrix prediction via graph convolution: A new perspective of passenger
  demand modeling,'' in \emph{Proceedings of the 25th ACM SIGKDD International
  Conference on Knowledge Discovery \& Data Mining}, ser. KDD '19.\hskip 1em
  plus 0.5em minus 0.4em\relax New York, NY, USA: Association for Computing
  Machinery, 2019, p. 1227–1235. [Online]. Available:
  \url{https://doi.org/10.1145/3292500.3330877}
\BIBentrySTDinterwordspacing

\bibitem{Wang2021PassengerMP}
Y.~Wang, H.~Yin, T.~Chen, C.~Liu, B.~Wang, T.~Wo, and J.~Xu, ``Passenger
  mobility prediction via representation learning for dynamic directed and
  weighted graph,'' \emph{ArXiv}, vol. abs/2101.00752, 2021.

\bibitem{Jin2020DeepMS}
G.~Jin, Z.~Xi, H.~Sha, Y.~Feng, and J.~Huang, ``Deep multi-view spatiotemporal
  virtual graph neural network for significant citywide ride-hailing demand
  prediction,'' \emph{ArXiv}, vol. abs/2007.15189, 2020.

\bibitem{Ke2019PredictingOR}
J.~Ke, X.~Qin, H.~Yang, Z.~Zheng, Z.~Zhu, and J.~Ye, ``Predicting
  origin-destination ride-sourcing demand with a spatio-temporal
  encoder-decoder residual multi-graph convolutional network,'' \emph{ArXiv},
  vol. abs/1910.09103, 2019.

\bibitem{Xu2019IncorporatingGA}
Y.~Xu and D.~Li, ``Incorporating graph attention and recurrent architectures
  for city-wide taxi demand prediction,'' \emph{ISPRS Int. J. Geo Inf.},
  vol.~8, p. 414, 2019.

\bibitem{8566163}
J.~Ke, H.~Yang, H.~Zheng, X.~Chen, Y.~Jia, P.~Gong, and J.~Ye, ``Hexagon-based
  convolutional neural network for supply-demand forecasting of ride-sourcing
  services,'' \emph{IEEE Transactions on Intelligent Transportation Systems},
  vol.~20, no.~11, pp. 4160--4173, 2019.

\bibitem{9101359}
H.~{Shi}, Q.~{Yao}, Q.~{Guo}, Y.~{Li}, L.~{Zhang}, J.~{Ye}, Y.~{Li}, and
  Y.~{Liu}, ``Predicting origin-destination flow via multi-perspective graph
  convolutional network,'' in \emph{2020 IEEE 36th International Conference on
  Data Engineering (ICDE)}, 2020, pp. 1818--1821.

\bibitem{Wang2020AnUC}
Y.~Wang, D.~Xu, P.~Peng, Q.~Xuan, and G.~jun Zhang, ``An urban commuters' od
  hybrid prediction method based on big gps data.'' \emph{Chaos}, vol. 30 9, p.
  093128, 2020.

\bibitem{Pian2020SpatialTemporalDG}
W.~Pian and Y.~Wu, ``Spatial-temporal dynamic graph attention networks for
  ride-hailing demand prediction,'' \emph{ArXiv}, vol. abs/2006.05905, 2020.

\bibitem{10.1145/2020408.2020579}
\BIBentryALTinterwordspacing
E.~Cho, S.~A. Myers, and J.~Leskovec, ``Friendship and mobility: User movement
  in location-based social networks,'' in \emph{Proceedings of the 17th ACM
  SIGKDD International Conference on Knowledge Discovery and Data Mining}, ser.
  KDD '11.\hskip 1em plus 0.5em minus 0.4em\relax New York, NY, USA:
  Association for Computing Machinery, 2011, p. 1082–1090. [Online].
  Available: \url{https://doi.org/10.1145/2020408.2020579}
\BIBentrySTDinterwordspacing

\bibitem{10.5555/3298239.3298479}
J.~Zhang, Y.~Zheng, and D.~Qi, ``Deep spatio-temporal residual networks for
  citywide crowd flows prediction,'' in \emph{Proceedings of the Thirty-First
  AAAI Conference on Artificial Intelligence}, ser. AAAI'17.\hskip 1em plus
  0.5em minus 0.4em\relax AAAI Press, 2017, p. 1655–1661.

\bibitem{10.5555/3295222.3295349}
A.~Vaswani, N.~Shazeer, N.~Parmar, J.~Uszkoreit, L.~Jones, A.~N. Gomez,
  u.~Kaiser, and I.~Polosukhin, ``Attention is all you need,'' in
  \emph{Proceedings of the 31st International Conference on Neural Information
  Processing Systems}, ser. NIPS'17.\hskip 1em plus 0.5em minus 0.4em\relax Red
  Hook, NY, USA: Curran Associates Inc., 2017, p. 6000–6010.

\bibitem{velickovic2018graph}
\BIBentryALTinterwordspacing
P.~Veli{\v{c}}kovi{\'{c}}, G.~Cucurull, A.~Casanova, A.~Romero, P.~Li{\`{o}},
  and Y.~Bengio, ``{Graph Attention Networks},'' \emph{International Conference
  on Learning Representations}, 2018. [Online]. Available:
  \url{https://openreview.net/forum?id=rJXMpikCZ}
\BIBentrySTDinterwordspacing

\bibitem{Zhang2018GaANGA}
J.~Zhang, X.~Shi, J.~Xie, H.~Ma, I.~King, and D.~Yeung, ``Gaan: Gated attention
  networks for learning on large and spatiotemporal graphs,'' in \emph{UAI},
  2018.

\bibitem{7410526}
R.~Girshick, ``Fast r-cnn,'' in \emph{2015 IEEE International Conference on
  Computer Vision (ICCV)}, 2015, pp. 1440--1448.

\bibitem{NEURIPS2019_9015}
\BIBentryALTinterwordspacing
A.~Paszke, S.~Gross, F.~Massa, A.~Lerer, J.~Bradbury, G.~Chanan, T.~Killeen,
  Z.~Lin, N.~Gimelshein, L.~Antiga, A.~Desmaison, A.~Kopf, E.~Yang, Z.~DeVito,
  M.~Raison, A.~Tejani, S.~Chilamkurthy, B.~Steiner, L.~Fang, J.~Bai, and
  S.~Chintala, ``Pytorch: An imperative style, high-performance deep learning
  library,'' in \emph{Advances in Neural Information Processing Systems 32},
  H.~Wallach, H.~Larochelle, A.~Beygelzimer, F.~d\textquotesingle
  Alch\'{e}-Buc, E.~Fox, and R.~Garnett, Eds.\hskip 1em plus 0.5em minus
  0.4em\relax Curran Associates, Inc., 2019, pp. 8024--8035. [Online].
  Available:
  \url{http://papers.neurips.cc/paper/9015-pytorch-an-imperative-style-high-performance-deep-learning-library.pdf}
\BIBentrySTDinterwordspacing

\bibitem{wang2019dgl}
M.~Wang, D.~Zheng, Z.~Ye, Q.~Gan, M.~Li, X.~Song, J.~Zhou, C.~Ma, L.~Yu,
  Y.~Gai, T.~Xiao, T.~He, G.~Karypis, J.~Li, and Z.~Zhang, ``Deep graph
  library: A graph-centric, highly-performant package for graph neural
  networks,'' \emph{arXiv preprint arXiv:1909.01315}, 2019.

\end{thebibliography}
% \nocite{*} % for every item

%%% About the authors %%%
\newpage
\phantomsection

\begin{IEEEbiography}[{\includegraphics[width=1in,height=1.25in,clip,keepaspectratio]{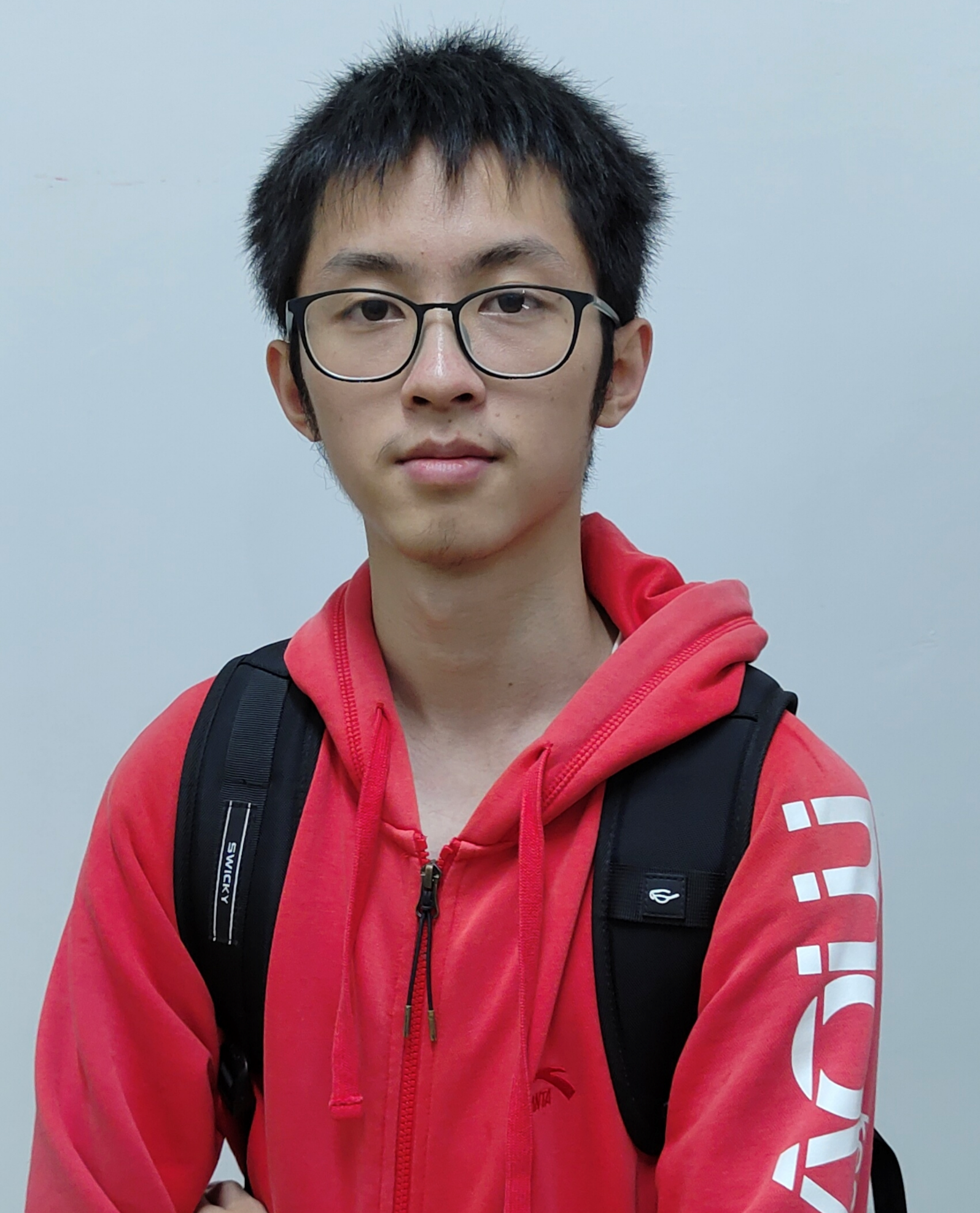}}]{Jingran Shen} received his B.Eng. degree in Computer Science and Technology from Southern University of Science and Technology (SUSTech), China in 2021. He is currently pursuing his M.Eng. in Computer Science at SUSTech. His research interests include deep learning, edge intelligence and cognitive science.
\end{IEEEbiography}

\begin{IEEEbiography}[{\includegraphics[width=1in,height=1.25in,clip,keepaspectratio]{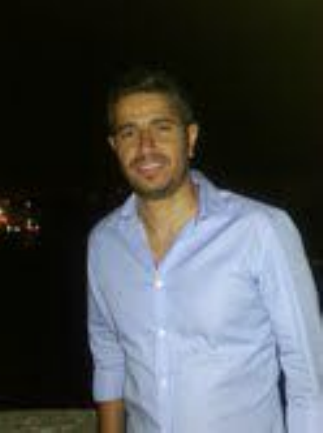}}]{Nikos Tziritas} received his Ph.D. degree from the University of Thessaly, Greece in 2011. He was an Associate Professor at the Shenzhen Institutes of Advanced Technology,  Chinese Academy of Sciences, China. In 2020 he joined the University of Thessaly as an Assistant Professor in the Department of Informatics and Telecommunications. His work has appeared in over 60  publications. He is the recipient of the Award for Excellence for Early Career Researchers in Scalable Computing from IEEE Technical Committee in Scalable Computing in 2016.
 \end{IEEEbiography}
 
\begin{IEEEbiography}[{\includegraphics[width=1in,height=1.25in,clip,keepaspectratio]{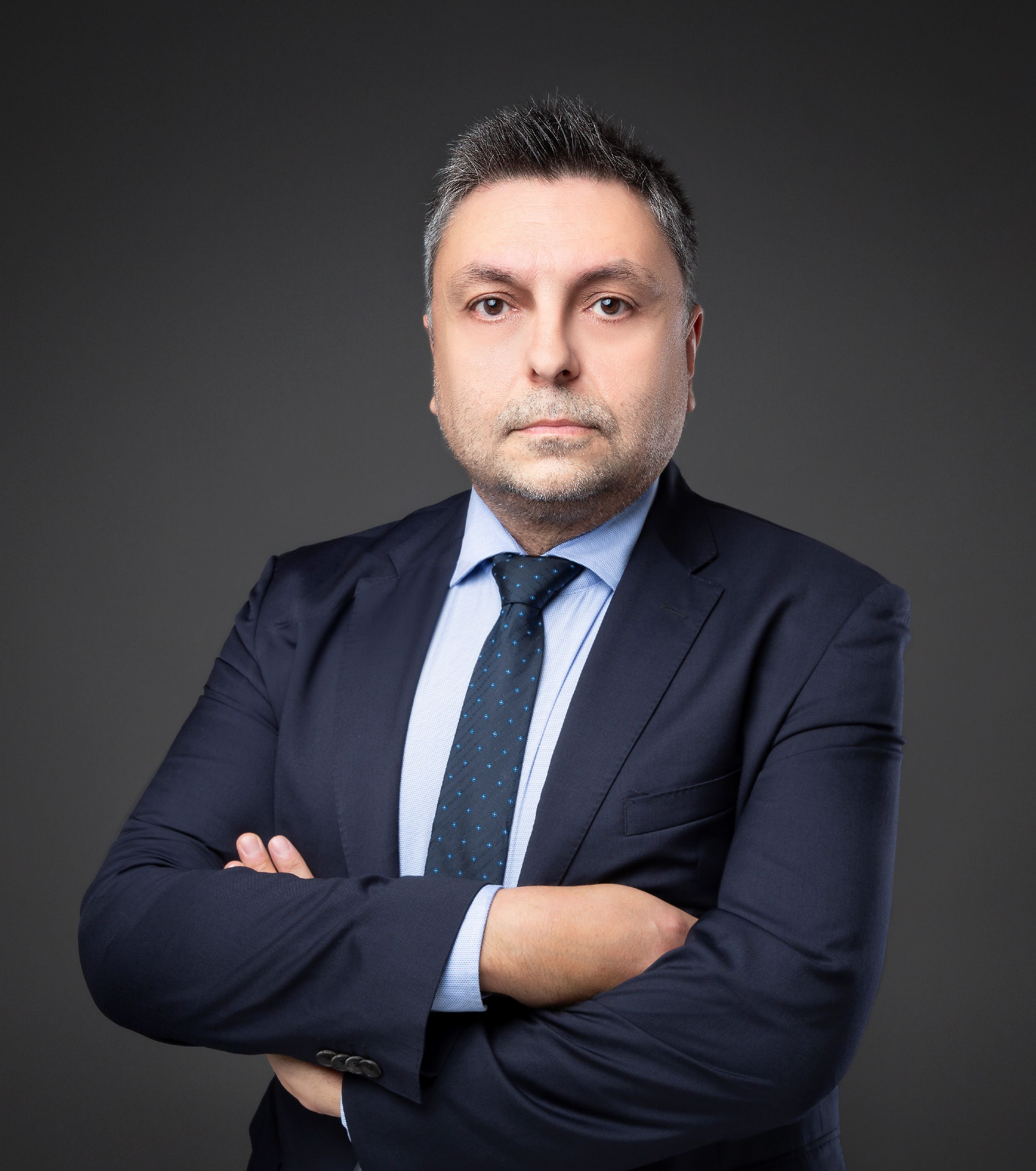}}]{Georgios Theodoropoulos} is currently a Chair Professor at the Department of Computer Science and Engineering at SUSTech in Shenzhen, China’s Silicon Valley. He joined SUSTech from Durham University, UK where he was the inaugural Executive Director of the Institute of Advanced Research Computing, a Chair Professor in Computer Engineering and the Head of the Innovative Computing Group at the School of Engineering and Computing Sciences. In the past, he was a Senior Research Scientist with IBM Research, held an Adjunct Chair at the Trinity College Dublin and senior faculty positions at the Nanyang Technological University, Singapore and the University of Birmingham, UK, where he was also founding Director of one of UK's National e-Science Centres of Excellence. He is a Chartered Engineer and holds a Ph.D. from the University of Manchester, UK. He is a Fellow of the World Academy of Art and Science.

\end{IEEEbiography}

\EOD
\end{document}